\documentclass{article} 
\usepackage{iclr2023_conference,times}

\usepackage{amsmath,amsfonts,bm}

\def\eqref#1{equation~\ref{#1}}

\def\1{\bm{1}}

\DeclareMathAlphabet{\mathsfit}{\encodingdefault}{\sfdefault}{m}{sl}
\SetMathAlphabet{\mathsfit}{bold}{\encodingdefault}{\sfdefault}{bx}{n}

\usepackage{url}
\usepackage{graphicx}
\usepackage{amsmath}
\usepackage{amssymb}
\usepackage{subcaption}
\usepackage{adjustbox}
\usepackage{color, colortbl}

\usepackage{booktabs}       
\usepackage{amsfonts}       
\usepackage{amsmath}
\usepackage{nicefrac}       
\usepackage{microtype}      
\usepackage{xcolor}         
\usepackage{multirow}
\usepackage{multicol}
\usepackage{xspace}
\usepackage{adjustbox}
\usepackage{import}
\usepackage{subcaption}
\usepackage{color, colortbl}
\usepackage{makecell}
\usepackage{wrapfig}
\usepackage{enumitem}
\usepackage{threeparttable}

\usepackage{color}
\usepackage{xcolor}
\definecolor{citecolor}{HTML}{0071bc}
\definecolor{mlpMixerColor}{RGB}{242,122,130}
\definecolor{concatColor}{RGB}{0,118,186}
\definecolor{dwconvColor}{RGB}{254,174,0}

\usepackage[pagebackref=true,breaklinks=true,letterpaper=true,urlcolor=citecolor,colorlinks,citecolor=citecolor,bookmarks=false]{hyperref}

\title{GPViT: A High Resolution Non-Hierarchical Vision Transformer with Group Propagation}

\author{Chenhongyi Yang\thanks{Equal Contribution}$~~$\textsuperscript{1} \quad Jiarui Xu$^*$\textsuperscript{2} \quad Shalini De Mello\textsuperscript{3} \quad Elliot J. Crowley\textsuperscript{1} \quad Xiaolong Wang\textsuperscript{2}\\
\textsuperscript{1}School of Engineering, University of Edinburgh \quad \textsuperscript{2}UC San Diego \quad \textsuperscript{3}NVIDIA}

\newcommand{\tablestyle}[2]{\setlength{\tabcolsep}{#1}\renewcommand{\arraystretch}{#2}\centering\large}

\iclrfinalcopy 
\begin{document}

\newcommand{\ourmethodFull}{Group Propagation Vision Transformer (GPViT)}
\newcommand{\ourmethod}{GPViT}
\newcommand{\ourblock}{GP Block}
\newcommand{\ourblockFull}{Group Propagation Block (GP Block)}
\newcommand{\ourblockFullNoAbbr}{Group Propagation Block}

\newcommand{\apbox}{AP$^{bb}$}
\newcommand{\apmask}{AP$^{mk}$}

\maketitle

\begin{abstract}
We present the \ourmethodFull: a novel non-hierarchical (i.e.\ non-pyramidal) transformer model designed for general visual recognition with high-resolution features. High-resolution features (or tokens) are a natural fit for tasks that involve perceiving fine-grained details such as detection and segmentation, but exchanging global information between these features is expensive in memory and computation because of the way self-attention scales. We provide a highly efficient alternative \ourblockFull~to exchange global information. In each \ourblock, features are first grouped together by a fixed number of learnable group tokens; we then perform~\emph{Group Propagation} where global information is exchanged between the grouped features; finally, global information in the updated grouped features is returned back to the image features through a transformer decoder. We evaluate \ourmethod~on a variety of visual recognition tasks including image classification, semantic segmentation, object detection, and instance segmentation. Our method achieves significant performance gains over previous works across all tasks, especially on tasks that require high-resolution outputs, for example, our \ourmethod-L3~outperforms Swin Transformer-B by 2.0 mIoU on ADE20K semantic segmentation with only half as many parameters. 

Project page: \url{chenhongyiyang.com/projects/GPViT/GPViT}
\end{abstract}

\begin{wrapfigure}{r}{0.55\textwidth}
  \vspace{-0.3in}
  \begin{center}
    \includegraphics[width=\linewidth]{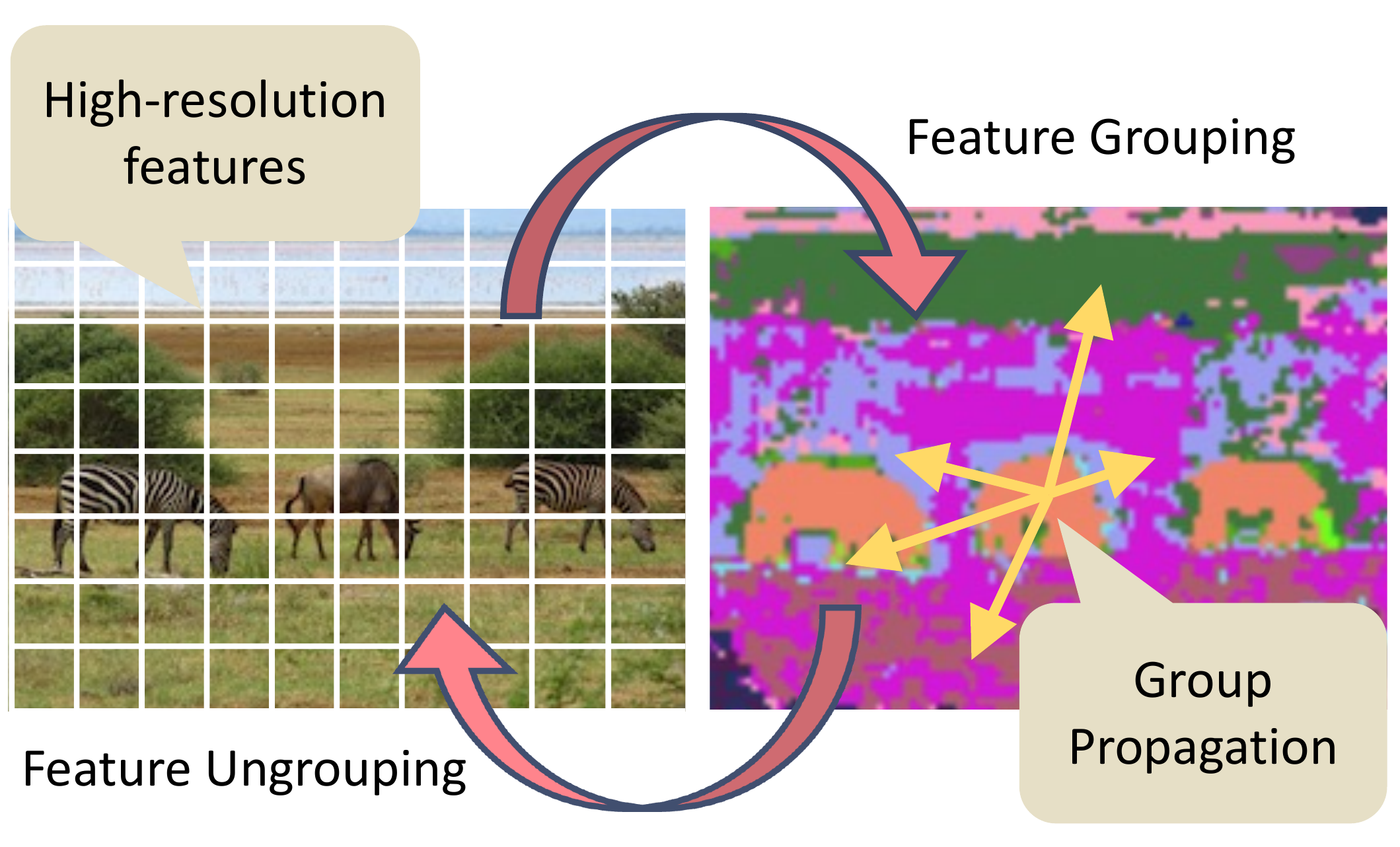}
  \end{center}
  \vspace{-0.2in}
    \caption{\small An illustration of our \ourblock. It groups image features into a fixed-size feature set. Then, global information is efficiently propagated between the grouped features. Finally, the grouped features are queried by the image features to transfer this global information into them.}
    \vspace{-0.1in}
    \label{fig:illustration}
\end{wrapfigure}

\section{Introduction}
\vspace{-0.1in}

Vision Transformer (ViT) architectures have achieved excellent results in general visual recognition tasks, outperforming ConvNets in many instances. In the original ViT architecture, image patches are passed through transformer encoder layers, each containing self-attention and MLP blocks. The spatial resolution of the image patches is constant throughout the network. Self-attention allows for information to be exchanged between patches across the whole image i.e.\ globally, however it is computationally expensive and does not place an emphasis on local information exchange between nearby patches, as a convolution would. Recent work has sought to build convolutional properties back into vision transformers~\citep{Swin_Liu_2021tq,CvT_Wu_2021tw,PVT_wang2021} through a hierarchical (pyramidal) architecture. This design reduces computational cost, and improves ViT performance on tasks such as detection and segmentation. 

Is this design necessary for structured prediction? It incorporates additional inductive biases~  {e.g.\ the assumption that nearby image tokens contains similar information}, which contrasts with the motivation for ViTs in the first place. A recent study~\citep{li2022exploring} demonstrates that a plain non-hierarchical ViT, a model that maintains the same feature resolution in all layers (non-pyramidal), can achieve comparable performance on object detection and segmentation tasks to a hierarchical counterpart. How do we go one step further and~\emph{surpass} this? One path would be to increase feature resolution (i.e.\ the number of image tokens). A plain ViT with more tokens would maintain high-resolution features throughout the network as there is no downsampling. This would facilitate fine-grained, detailed outputs ideal for tasks such as object detection and segmentation. It also simplifies the design for downstream applications, removing the need to find a way to combine different scales of features in a hierarchical ViT. However, this brings new challenges in terms of computation. Self-attention has quadratic complexity in the number of image tokens. Doubling feature resolution (i.e.\ quadrupling the number of tokens) would lead to a $16\times$ increase in compute. How do we maintain global information exchange between image tokens without this huge increase in computational cost?

In this paper, we propose the \textbf{Group Propagation Vision Transformer (GPViT)}: a non-hierarchical ViT which uses high resolution features throughout, and allows for efficient global information exchange between image tokens. We design a novel \ourblockFull~ for use in plain ViTs. Figure~\ref{fig:illustration} provides a high-level illustration of how this block works. In detail, we use learnable group tokens and the cross-attention operation to group a large number of high-resolution image features into a fixed number of grouped features. Intuitively, we can view each group as a cluster of patches representing the same semantic concept. We then use an MLPMixer~\citep{tolstikhin2021mlp} module to update the grouped features and propagate global information among them. This process allows information exchange at a low computational cost, as the number of groups is much smaller than the number of image tokens. Finally, we ungroup the grouped features using another cross-attention operation where the updated grouped features act as key and value pairs, and are queried by the image token features. This updates the high resolution image token features with the group-propagated information. The \ourblock~only has a linear complexity in the number of image tokens, which allows it to scale better than ordinary self-attention. This block is the foundation of our simple non-hierarchical vision transformer architecture for general visual recognition. 

\begin{wrapfigure}{r}{0.55\textwidth}
  \vspace{-0.2in}
  \begin{center}
    \includegraphics[width=\linewidth]{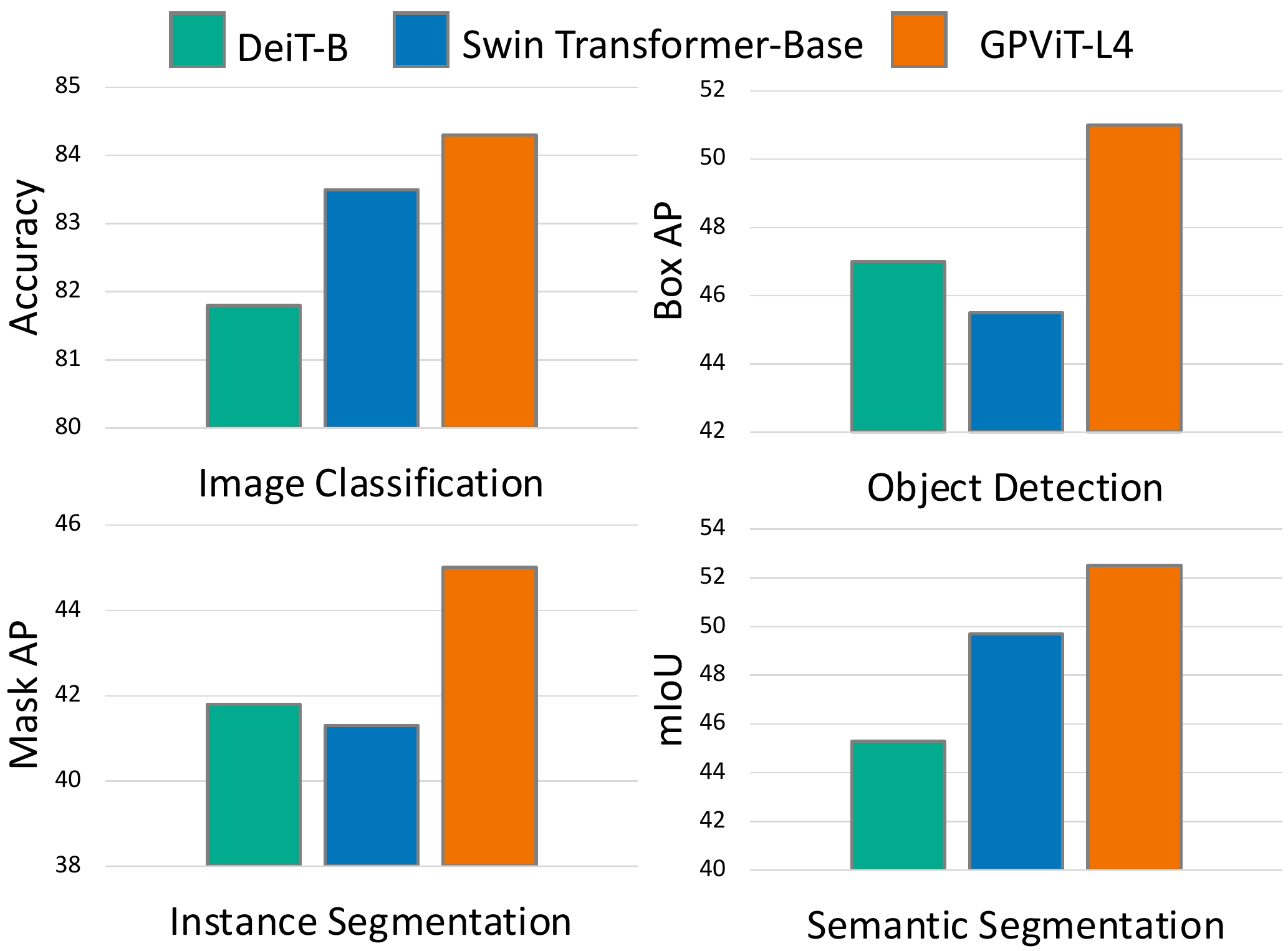}
  \end{center}
  \vspace{-0.2in}
    \caption{{\small {A comparison on four visual recognition tasks between \ourmethod~and the non-hierarchical DeiT~\citep{DeiT_touvron2020} and the hierarchical Swin Transformer~\citep{Swin_Liu_2021tq}.}}}
    \vspace{-0.1in}
    \label{fig:performance}
\end{wrapfigure}
We conduct experiments on multiple visual recognition tasks including image classification, object detection, instance segmentation, and semantic segmentation. We show significant improvements over previous approaches, including hierarchical vision transformers, under the same model size in all tasks. The performance gain is especially large for object detection and segmentation. For example, in Figure~\ref{fig:performance}, we show \ourmethod's advantage over the non-hierarchical DeiT~\citep{DeiT_touvron2020} and hierarchical Swin Transformer~\citep{Swin_Liu_2021tq} on those recognition tasks. In addition, our smallest model \ourmethod-L1 can outperform the Swin Transformer-B~\citep{Swin_Liu_2021tq} by 2.6 AP$^{bb}$ and 1.4$^{mk}$ in COCO Mask R-CNN~\citep{MaskRCNN_He_2017_ICCV} object detection and instance segmentation with only 30\% as many parameters, and \ourmethod-L2 outperforms Swin Transformer-B by 0.5 mIoU on UperNet~\citep{xiao2018unified} ADE20K semantic segmentation also with only 40\%  as many parameters. 

\vspace{-0.1in}
\section{Related Work}

\vspace{-0.1in}
\textbf{Vision Transformers.}
Vision Transformers have shown great success in visual recognition. They have fewer inductive biases,  {e.g.\ translation invariance, scale-invariance, and feature locality~\citep{xu2021vitae}} than ConvNets and can better capture long-range relationships between image pixels. In the original ViT architecture~\citep{ViT_dosovitskiy2021an,DeiT_touvron2020}, images are split into patches and are transformed into tokens that are passed through the encoder of a transformer~\citep{Transformer_NIPS2017_Vaswani}. Based on this framework, LeViT~\citep{LeViT_BenGraham_2021vh} achieves a significant performance improvement over ViT by combining convolutional and transformer encoder layers. An important  development in ViT architectures is the incorporation of a hierarchical feature pyramid structure, as typically seen in ConvNets~\citep{PVT_wang2021,Swin_Liu_2021tq,xu2021co,CvT_Wu_2021tw,fan2021multiscale}. For example, ~\citet{Swin_Liu_2021tq} propose a shifted windowing scheme to efficiently propagate feature information in the hierarchical ViT. Such a pyramid architecture provides multi-scale features for a wide range of visual recognition tasks. Following this line of research, recent work has studied the use of hierarchical features in ViTs~\citep{stunned,CMT,mvitv2, cswin, gcvit, chen2022regionvit, ConViTdAscoli_2021vz, MPViT}. For example, ~\citet{stunned} introduce using multi-resolution features as attention keys and values to make the model learn better multi-scale information. While this is encouraging, it introduces extra complexity in the downstream model's design on how to utilize the multi-scale features effectively. Recently,~\citet{li2022exploring} revisited the plain non-hierarchical ViT for visual recognition; using such a model simplifies the use of features and better decouples the pre-training and downstream stages of model design. Our work extends on this as we examine how to efficiently increase the feature resolution in a non-hierarchical ViT.

\textbf{Attention Mechanisms in ViTs.} 
A bottleneck when using high resolution features in ViTs is the quadratic complexity in the computation of the self-attention layer. To tackle this challenge, several local attention mechanisms have been proposed~\citep{Swin_Liu_2021tq,huang2019ccnet,cswin,xu2021co, zhang2022nested, han2021transformer} to allow each image token to attend to local region instead of the whole image. However, using only local attention hinders a model's ability to exchange information globally. To counter this problem, RegionViT~\citep{chen2022regionvit} and GCViT~\citep{gcvit} first down-sample their feature maps and exchange global information between the down-sampled features, before using self-attention to transfer information between the original image features and the down-sampled features. This is similar in spirit to our \ourblock. However, unlike RegionViT and GCViT, in a \ourblock~the grouped features are not constrained to a particular rectangular region, but can correspond to any shape or even entirely disconnected image parts.  {There is recent work using transformer decoder layers with cross-attention between visual tokens and learnable tokens~\citep{detr,cheng2021mask2former,jaegle2021perceiver,hudson2021generative}, however, there are three fundamental differences between these and ours: (i) Each of our GP blocks operates as an `encoder-decoder' architecture with two rounds of cross-attention between visual tokens and group tokens: the first round groups the visual tokens for group propagation, and the second round ungroups the updated groups back into visual tokens; (ii) The underlying functionality is different: GP blocks facilitate more efficient global information propagation throughout the ViT, while previous work applies the decoder to obtain the final results for inference (e.g bounding boxes, or masks in~\cite{detr,cheng2021mask2former}); (iii) The GP block is a general module that can be insert into any layer of the ViT, while previous work utilizes the decoder only in the end of the network. 
} 

\textbf{High-Resolution Visual Recognition.} 
Previous work~\citep{wang2020deep, HigherHRNet_Cheng_2020_CVPR} has shown that high-resolution images and features are beneficial to visual recognition tasks, especially to those requiring the perception of fine-grained image details, for example, semantic segmentation~\citep{wang2020deep}, pose-estimation~\citep{sun2019deep}, and small object detection~\citep{Yang_2022_CVPR}. For example, HRNet~\citep{wang2020deep} introduces a high-resolution ConvNet backbone. It maintains a high-resolution branch and exchanges information between different resolutions of features with interpolation and strided convolutions. Inspired by this work, HRFormer~\citep{yuan2021hrformer} and HRViT~\citep{gu2021hrvit} replace the convolutions in HRNet with self-attention blocks. \ourmethod~is even simpler: it maintains single-scale and high-resolution feature maps without requiring any cross-resolution information to be maintained. 

\textbf{Object-Centric Representation.} 
Our idea of performing information propagation among grouped regions is related to object-centric representation learning~\citep{wang2018videos,kipf2016semi,watters2017visual, qi2020learning,locatello2020object, kipf2021conditional, elsayed2022savi++, xu2022groupvit}. For example,~\citet{locatello2020object} proposes slot-attention, which allows automatic discovery of object segments via a self-supervised reconstruction objective. Instead of using reconstruction,~\citet{xu2022groupvit} utilizes language as an alternative signal for object segmentation discovery and shows it can be directly transferred to semantic segmentation in a zero-shot manner. All the above work extract object-centric features for downstream applications, while our work inserts this object-centric information propagation mechanism as a building block inside ViTs to compute high-resolution representations more efficiently and improve high-resolution features. In this respect, our work is related to~\cite{li2018beyond} where the graph convolution operations are inserted into ConvNets for better spatial reasoning.

\vspace{-0.1in}
\section{Method}
\begin{figure*}[!t]
    \centering
    \includegraphics[width=\textwidth]{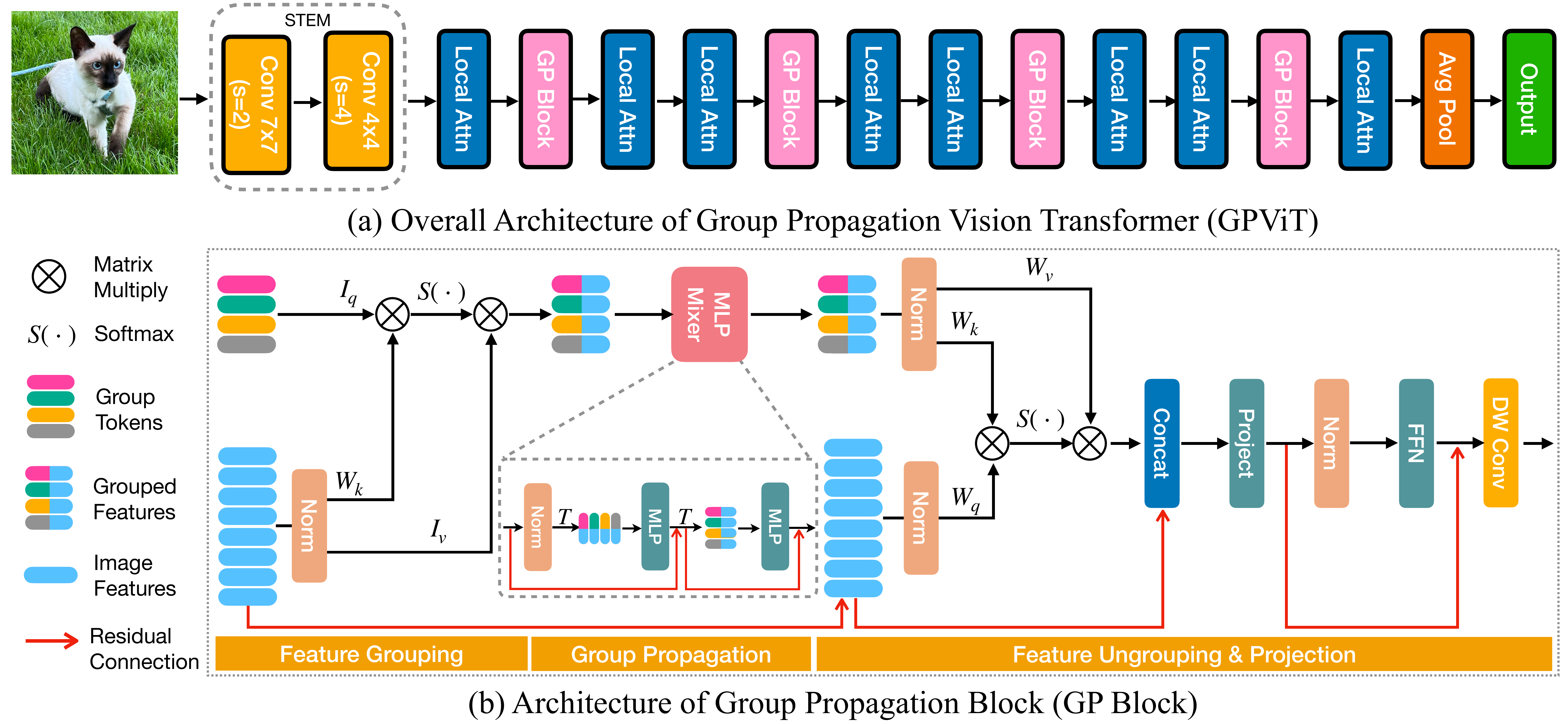}
\vspace{-0.2in}
    \captionsetup{font={small}}
\caption{(a). \ourmethod~architecture: The input image is first fed to a convolutional stem that downsamples by a factor of 8. Each pixel of this downsampled image is treated as a high resolution image token or~\emph{feature}, and positional embeddings are added (not shown in figure). These features are fed into our transformer, which consists of 12 encoder layers. 8 of these use local attention, and 4 use our proposed \ourblock~to propagate global information between image features. (b). \ourblock: Image features are grouped using a fixed number of learnable group tokens. An MLPMixer module is then used to exchange global information and update the grouped features. Next, the grouped features are queried by, and concatenated with the image features to transfer global information to every image feature. Finally the updated image features are transformed by a feed-forward network to produce the output.}
\label{fig:pipeline}
\end{figure*}

\vspace{-0.1in}
We present the overall architecture of our Group Propagation Vision Transformer (GPViT) in Figure~\ref{fig:pipeline} (a). \ourmethod~is designed for general high-resolution visual recognition. For stable training, we first feed the input image into a down-sampling convolutional stem to generate image features (also known as image tokens), as in~\cite{ViT_dosovitskiy2021an,Swin_Liu_2021tq}. In \ourmethod~we downsample by a factor of 8 by default. The features are therefore higher resolution than in the original ViT where the factor is 16. Unlike most recently proposed methods~\citep{Swin_Liu_2021tq, mvitv2} that adopt a pyramid structure to generate features in multiple resolutions, we keep the features at a high resolution without any down-sampling. 

After combining the initial image features with positional embeddings~\citep{Transformer_NIPS2017_Vaswani}, we feed them into the core \ourmethod~architecture. We replace the original self-attention block in ViT with local attention to avoid the quadratic complexity of self-attention. However, stacking local attention blocks alone does not allow for long-range information exchange between patches and therefore is harmful to performance. To counter this problem, we propose the \ourblockFull---which we describe in full in Section~\ref{sec:gpblock}---to efficiently propagate global information across the whole image. In our implementation, we use a mixture of~\ourblock s and local attention layers to form our \ourmethod~and keep the overall depth unchanged. Lastly, we average the final features to get the model's output.

\subsection{Group Propagation Block}
\label{sec:gpblock}

Our key technical contribution is the GP block, which efficiently exchanges global information between each image patch with a linear complexity. We visualize the structure of the GP block in Figure~\ref{fig:pipeline} (b). It has a bottleneck structure and comprises of three stages, namely, \textit{Feature Grouping}, \textit{Group Propagation}, and \textit{Feature Ungrouping}. In the first stage the image features are grouped, then in the second stage global information is propagated between the grouped features, and in the last stage, this global information is transferred back to the image features. 

\textbf{Feature Grouping.} The input to a \ourblock~is a matrix of image features $X \in \mathbb{R}^{N \times C}$   {(The blue tokens in Figure~\ref{fig:pipeline} (b))} where $N$ is the total number of image features (or image tokens) and $C$ is the dimensionality of each feature vector. We use $M$~learnable group tokens stored in a matrix $G \in \mathbb{R}^{M\times C}$   {(the multi-colored tokens in Figure~\ref{fig:pipeline} (b))} where the group number $M$ is a model hyper-parameter. Grouping is performed using a simplified multi-head attention operation~\citep{Transformer_NIPS2017_Vaswani}, which gives us grouped features $Y \in \mathbb{R}^{M\times C}$ (the half-and-half tokens in Figure~\ref{fig:pipeline} (b)):
\begin{align}
    &\text{Attention}(Q, K, V) = \text{Softmax}(\frac{QK^T}{\sqrt{d}})V, \\
    &Y = \text{Concat}_{\{h\}}\bigl(\text{Attention}(W_{h}^{Q}G_h, W_{h}^{K}X_h, W_{h}^{V}X_h)\bigr),
\end{align}
where $d$ is the channel number, $h$ is the head index, and $W_{h}^{\{Q,K,V\}}$ are projection matrices for the query, key, and values, respectively in the attention operation. We remove the feature projection layers after the concatenation operation and set $W_{h}^Q$ and $W_{h}^{V}$ to be identity matrix. Therefore, the grouped features are simply the weighted sum of image features at each head where the weights are computed by the attention operation.

\textbf{Group Propagation.} After acquiring the grouped features, we can update and propagate global information between them. We use an MLPMixer~\citep{tolstikhin2021mlp}   {(Equation~\ref{eq:mlpmixer}; the \textcolor{mlpMixerColor}{red} box in Figure~\ref{fig:pipeline} (b))} to achieve this, as MLPMixer provides a good trade-off between model parameters, FLOPs, and model accuracy. MLPMixer requires a fixed-sized input, which is compatible with our fixed number of groups. Specifically, our MLPMixer contains two consecutive MLPs. Recall that $Y \in \mathbb{R}^{M\times C}$ contains the grouped features from the first \textit{Feature Grouping} stage. We can update these features to $\tilde{Y} \in \mathbb{R}^{M\times C}$ with the MLPMixer by computing:
\begin{align}
\label{eq:mlpmixer}
    & Y' = Y + \text{MLP}_1(\text{LayerNorm}(Y)^T))^T,\\ & \tilde{Y} = Y' + \text{MLP}_2(\text{LayerNorm}(Y'))),
\end{align}
where the first MLP is used for mixing information between each group, and the second is used to mix channel-wise information.

\textbf{Feature Ungrouping.} After updating the grouped features, we can return global information to the image features through a \textit{Feature Ungrouping} process. Specifically, the features are ungrouped using a transformer decoder layer where grouped features are queried by the image features.
\begin{align}
\label{eqa:ungroup1}
    &U = \text{Concat}_{\{h\}}\bigl(\text{Attention}(\tilde{W}_{h}^{Q}X_h, \tilde{W}_{h}^{K}\tilde{Y}_h, \tilde{W}_{h}^{V}\tilde{Y}_h)\bigr), \\
    Z' = W_{proj} &* \text{Concat}(U, X),~~~~~Z'' = Z' + \text{FFN}(Z'),~~~~~Z = \text{DWConv}(Z''), \label{eqa:ungroup2}
\end{align}
where $\tilde{W}_{h}^{\{Q,K,V\}}$ are the projection matrices in the attention operation, $W_{proj}$ is a linear matrix that projects concatenated features $Z'$ to the same dimension as image features $X$, $\text{FFN}$ is a feed-forward network, and $\text{DWConv}$ is a depth-wise convolution layer. We modify the original transformer decoder layer by replacing the first residual connection with a concatenation operation   {(Equation~\ref{eqa:ungroup1}; the \textcolor{concatColor}{blue} box in Figure~\ref{fig:pipeline} (b))}, and move the feature projection layer after this to transform the feature to the original dimension. We find this modification benefits the downstream tasks in different sizes of models. We take inspiration from~\cite{stunned} and add a depth-wise convolution at the end of the \ourblock~to improve the locality property of the features   {(Equation~\ref{eqa:ungroup2}; the \textcolor{dwconvColor}{yellow} box in Figure~\ref{fig:pipeline} (b))}. Finally, a \ourblock~outputs $Z$  as its final output.

\setlength\intextsep{0pt}
\begin{table}[!h]
    \centering
    \vspace{3mm}
    \caption{\small {\ourmethod~architecture variants. }}
    \vspace{-3mm}
    \begin{adjustbox}{max width=0.45\linewidth}
    \begin{tabular}{lccc}
        \toprule
        Model & Channels & Param (M) & FLOPs (G)\\
        \midrule
        \ourmethod-L1 & 216 & 9.3 & 5.8 \\
        \ourmethod-L2 & 348 & 23.6 & 15.0 \\
        \ourmethod-L3 & 432 & 36.2 & 22.9 \\
             {\ourmethod-L4} &      {624} &      {75.4} &      {48.2} \\
        \bottomrule
    \end{tabular}
    \end{adjustbox}
    \label{table:architecture}
    \vspace{1mm}
\end{table}

\subsection{Architecture Variants of \ourmethod}
In this paper we study four variants of the proposed \ourmethod. We present their architectural details in Table~\ref{table:architecture}. These four variants largely differ in the number of feature channels used (i.e.\ the model width). We use the recently proposed LePE attention~\citep{cswin} as local attention by default. The FLOPs are counted using 224$\times$224 inputs. Please refer to Section~\ref{sec:modelDetails} in our Appendix for detailed architectural hyper-parameters and training recipes for these variants.

\subsection{Computational costs of hierarchical and non-hierarchical ViTs.}
\label{sec:computation}

\begin{wrapfigure}{r}{0.6\textwidth}
  \vspace{0.1in}
  \begin{center}
    \includegraphics[width=\linewidth]{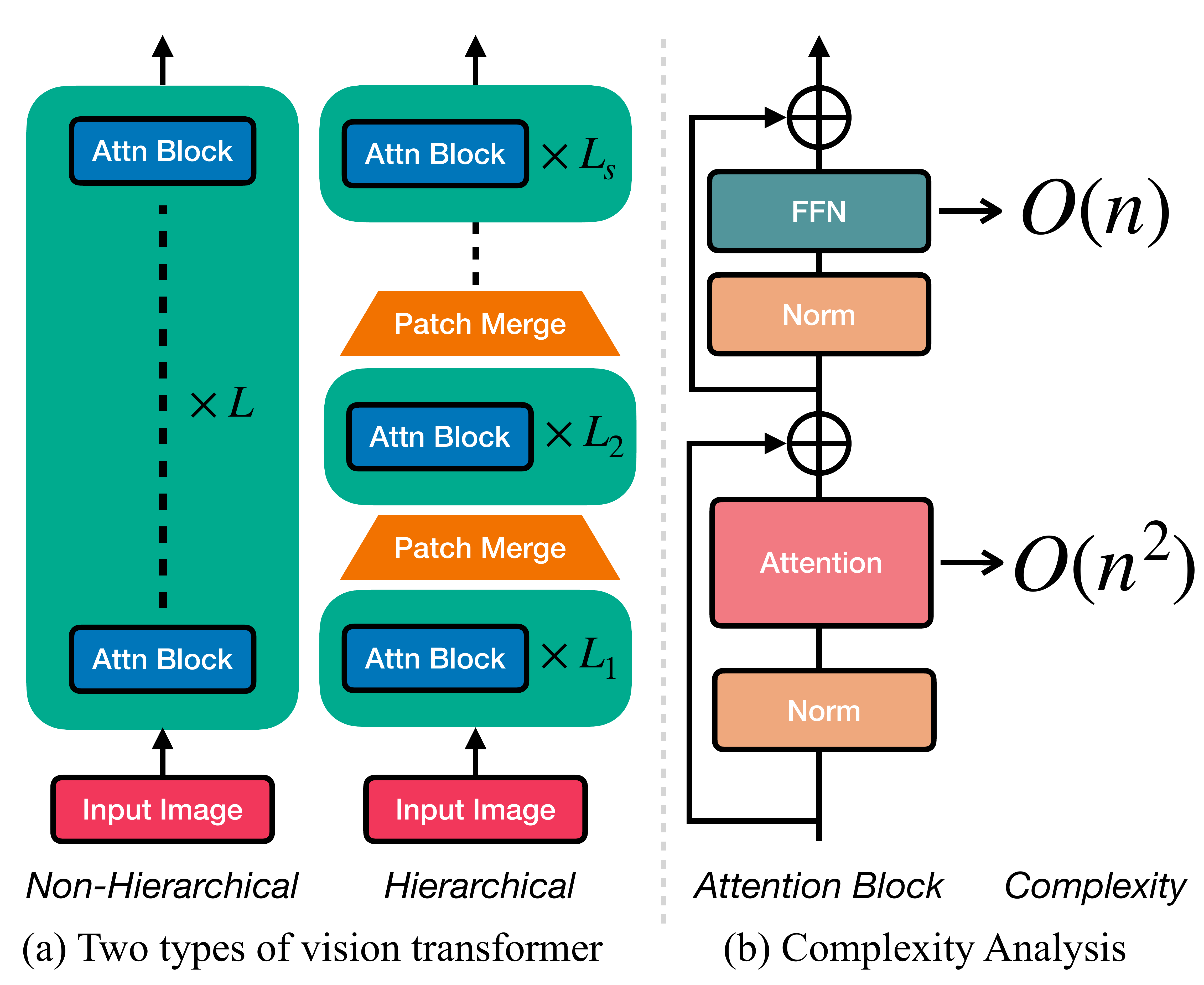}
  \end{center}
  \vspace{-0.2in}
    \caption{{\small {(a). Structure comparison between non-hierarchical and hierarchical ViTs. (b). The computation cost of an attention block.}}}
    \vspace{0.1in}
    \label{fig:stuctureComplexity}
\end{wrapfigure}
We visualize both the non-hierarchical and hierarchical ViT in Figure~\ref{fig:stuctureComplexity} (a), where the non-hierarchical ViT simply stacks attention blocks and the hierarchical ViT divides the network into several stages and down-samples the feature map at each stage. Naturally, with the same resolution input, the non-hierarchical ViT will have a higher computation cost. The cost is divided into two parts as shown in Figure~\ref{fig:stuctureComplexity} (b): the self-attention module and the FFN module. Our GPViT largely reduces the computation of information propagation by using our GP Block instead of self-attention. However, the cost of the FFN stays high for high resolution image features. Therefore we will expect higher FLOPs from GPViT compared to a hierarchical ViT given similar model parameters. However, we believe non-hierarchical ViTs are still a direction worthy of exploration given their simplicity in extracting high-resolution features and the removal of the need to study the design of efficient downstream models that utilize multi-scale features as required for a hierarchical ViT. This helps to maintain the independence of the model's pre-training and fine-tuning designs~\citep{li2022exploring}. In our experiments, we show that our GPViT can achieve better detection and segmentation performance compared to state-of-the-art hierarchical ViTs with similar FLOP counts.

\vspace{-0.1in}
\section{Experiments}
\vspace{-0.1in}


\subsection{ImageNet-1K Classification}
\vspace{-0.1in}
\label{sec:expIn1k}

\textbf{Setting:} To ensure a fair comparison with previous work, we largely follow the training recipe of Swin Transformer~\citep{Swin_Liu_2021tq}. We build models using the MMClassification~\citep{2020mmclassification} toolkit. The models are trained for 300 epochs with a batch size of 2048 using the AdamW optimizer with a weight decay of 0.05 and a peak learning rate of 0.002. A cosine learning rate schedule is used to gradually decrease the learning rate. We use the data augmentations from \cite{Swin_Liu_2021tq}; these include Mixup~\citep{zhang2017mixup}, Cutmix~\citep{yun2019cutmix}, Random erasing~\citep{zhong2020random} and Rand augment~\citep{cubuk2020randaugment}.

\setlength\intextsep{0pt}
\begin{wraptable}{r}{0.45\linewidth}
    \centering
    \caption{\small Comparison between \ourmethod~and the recent proposed models on ImageNet-1K.}
    \label{table:in1kSmall}
    \vspace{-3mm}
    \begin{adjustbox}{max width=\linewidth}
\tablestyle{2pt}{0.9}
\begin{tabular}{l@{\hskip -5pt}ccc}
\toprule
      &  Params & FLOPs  & Top-1   \\
Model &  (M) & (G)  & Acc   \\
\midrule
\multicolumn{4}{l}{Hierarchical} \\
\midrule
Swin-T~\citep{Swin_Liu_2021tq} &    29.0 &     4.5 &  81.3 \\
Swin-B~\citep{Swin_Liu_2021tq} &    88 &    15.4 &  83.5 \\
RegionViT-S~\citep{chen2022regionvit} &    30.6 &     5.3 &  82.6 \\
RegionViT-B~\citep{chen2022regionvit} &    72.7 &     13.0 &  83.2 \\
DW-T~\citep{dwvit} & 30.0 &   5.2 & 82.0 \\
DW-B~\citep{dwvit} & 91.0 &   17.0 & 83.8 \\
\midrule
\multicolumn{4}{l}{Non-hierarchical} \\
\midrule
DeiT-S~\citep{DeiT_touvron2020} &    22.1 &     4.6 &  79.9 \\
DeiT-B~\citep{DeiT_touvron2020} &    86 &     16.8 &  81.8 \\
ConViT-S~\citep{ConViTdAscoli_2021vz} &    27.0 &     5.4 &  81.3 \\
ConViT-B~\citep{ConViTdAscoli_2021vz} &    86.0 &     17.0 &  82.4\\
\midrule
\ourmethod-L1 & 9.3 &   5.8 & 80.5 \\
\ourmethod-L2 & 23.8 &   15.0 & 83.4 \\
\ourmethod-L3 & 36.2 &   22.9 & 84.1 \\
   {\ourmethod-L4} &    {75.4} &      {48.2} &    {84.3} \\
\bottomrule
\end{tabular}
\end{adjustbox}
\vspace{-1mm}

\end{wraptable}

We compare \ourmethod~with hierarchical and non-hierarchical vision transformers on the ImageNet-1K classification task and report the results in Table~\ref{table:in1kSmall}. As shown in the table, because of our high-resolution design and effective global information propagation via the grouping mechanism, our \ourmethod~outperforms outperforms the non-hierarchical baseline DeiT~\citep{DeiT_touvron2020}. In addition, \ourmethod~also outperforms Swin Transformer~\citep{Swin_Liu_2021tq} and two recently proposed hierarchical counterparts RegionViT~\citep{chen2022regionvit} and DWViT~\citep{dwvit}. This result showcases the potential of non-hierarchical vision transformers and suggests that the hierarchical design inherited from the ConvNet era is not necessary for obtaining a high-performing visual recognition model. This corroborates the work of~\cite{li2022exploring}. That said, we do note that the FLOPs of our models are higher than most alternatives for a similar parameter count. However, for a similar FLOP count we observe that \ourmethod~can achieve a comparable top-1 accuracy, but with many fewer parameters than the alternatives. For example, \ourmethod-L2 (15.0 G) has similar FLOPs to the Swin Transformer-B (15.4 G) and ShiftViT-B (15.6 G), but it achieves a similar accuracy with significantly fewer parameters (23.8 M v.s. 88 M and 89 M).

\subsection{COCO object detection and instance segmentation}

\textbf{Setting:} We follow~\citet{vitadapter} to use Mask R-CNN and RetinaNet models for the COCO object detection and instance segmentation tasks. We use ViTAdapter~\citep{vitadapter} to generate multi-scale features as FPN inputs and evaluate the model for both 1$\times$ and 3$\times$ training schedules.

\begin{table}[!t]
\centering
\caption{\small Mask R-CNN object detection and instance segmentation on MS COCO \textit{mini-val} using 1$\times$ and 3$\times$ (or longer) + MS schedule.}
\label{table:maskrcnnSmall}
\vspace{-3mm}
\begin{adjustbox}{max width=.65\linewidth}
\tablestyle{2pt}{0.95}
\begin{tabular}{lcc|cc|cc}
\toprule
         &      Params      &    FLOPs       & \multicolumn{2}{c|}{1$\times$} & \multicolumn{2}{c}{3$\times$ or more} \\
Backbone &  (M) &  (G) & $AP^{bb}$  & $AP^{mk}$ & $AP^{bb}$ & $AP^{mk}$  \\
\midrule
\multicolumn{7}{l}{Hierarchical} \\
\midrule
RegionViT-B~\citep{chen2022regionvit} & 92 & 287 & 44.2 & 40.8 & 47.6 & 43.4  \\
Swin-B~\citep{Swin_Liu_2021tq} & 107 & 496 & 45.5 & 41.3 & - & - \\
DaViT-Base~\citep{DAViT} & 107 & 491 & 48.2 & 43.3 & 49.9 & 44.6\\
DW-B~\citep{dwvit} & 111 & 505 & - & - & 49.2 & 44.0 \\
CSwin-B~\citep{cswin} &    97 &    526 &  48.7 & 43.9 & 50.8 & 44.9 \\
MPViT-B~\citep{MPViT} &    95 &    503 &  - & - & 49.5 & 44.5 \\
\midrule
\multicolumn{7}{l}{Non-hierarchical} \\
\midrule
ViT-Adapter-B~\citep{vitadapter} & 102 & - & 47.0 & 41.8 & 49.6 & 43.6 \\
ViTDet-SUP\textcolor{red}{$^*$}~\citep{li2021benchmarking} & 111 & 800 & - & - & 47.6 & 42.4 \\
ViTDet-MAE\textcolor{red}{$^*$}~\citep{li2021benchmarking} & 111 & 800 & - & - & 51.6 & 45.2 \\
\midrule
\ourmethod-L1 & 33 & 457 & 48.1 & 42.7 & 50.2 & 44.3 \\
\ourmethod-L2 & 50 & 690 & 49.9 & 43.9 & 51.4 & 45.1 \\
\ourmethod-L3 & 64 & 884 & 50.4 & 44.4 & 51.6 & 45.2 \\
{\ourmethod-L4} &     {109} &     {1489} &     {51.0} &     {45.0} &     {52.1} &     {45.7} \\
\bottomrule
\multicolumn{7}{l}{\footnotesize \textcolor{red}{$^*$}: ViTDet~\citep{li2021benchmarking} models were trained for 100 epochs with advanced regularisation techniques.}

\end{tabular}
\end{adjustbox}
\vspace{-6mm}

\end{table}

\setlength\intextsep{0pt}
\begin{wraptable}{r}{0.46\linewidth}
\centering
\caption{\small RetinaNet object detection on MS COCO \textit{mini-val} with 1$\times$ and 3$\times$ +MS schedule.}
\label{table:retinanetSmall}
\vspace{-3mm}
\begin{adjustbox}{max width=\linewidth}
\tablestyle{2pt}{0.9}
\begin{tabular}{l@{\hskip -5pt}cc|c|c}
\toprule
         &       Params      &     FLOPs       & 1$\times$  & 3$\times$ \\
Backbone &   (M) &   (G) & $AP^{bb}$  & $AP^{bb}$ \\
\midrule
\multicolumn{5}{l}{Hierarchical} \\
\midrule
PVT-L~\citep{PVT_wang2021} & 71  & 345 & 42.6 & - \\
PVTv2-B5~\citep{pvtv2} & 91  & 335 & 46.1 & - \\
Swin-B~\citep{Swin_Liu_2021tq} & 98 & 477 & 44.7 & - \\
RegionViT-B~\citep{chen2022regionvit} & 83 & 308 & 43.3 & 46.1  \\
MPViT-B~\citep{MPViT} & 95 & 503 & - & 48.3  \\
DaViT-Base~\citep{DAViT} & 103 & 471 & 46.7 & 48.7 \\
\midrule
\multicolumn{5}{l}{Non-hierarchical} \\
\midrule
\ourmethod-L1 & 21 & 317 & 45.8 & 48.1 \\
\ourmethod-L2 & 37 & 542 & 48.0 & 49.0 \\
\ourmethod-L3 & 52 & 731 & 48.3 & 49.4 \\
{\ourmethod-L4} &   {96} &   {1319} &   {48.7} &   {49.8} \\
\bottomrule 
\end{tabular}
\end{adjustbox}

\end{wraptable}
\textbf{Results:} We compare \ourmethod~to state-of-the-art backbones, all pre-trained on ImageNet-1K. We report the results in Table~\ref{table:maskrcnnSmall} and Table~\ref{table:retinanetSmall}. For competing methods we report the performance of their largest-sized models. For both detectors our \ourmethod~is able to surpass the other backbones by a large margin for a similar parameter count. With Mask R-CNN (Table~\ref{table:maskrcnnSmall}), our smallest \ourmethod-L1 surpasses its Swin Transformer-B~\citep{Swin_Liu_2021tq} counterpart by 2.6 AP$^{bb}$ and 1.4 AP$^{mk}$ for the 1$\times$ training schedule with fewer FLOPs and only 30\% as many parameters. When comparing with models that are also equipped with ViTAdapter~\citep{vitadapter}, we observe that \ourmethod~achieves a better AP with fewer parameters, e.g.\ our smallest \ourmethod-L1 outperforms ViT-Adapter-B in both training schedules. These results showcase \ourmethod's effectiveness at extracting good regional features for object detection and instance segmentation. A similar conclusion can be drawn from the single-stage RetinaNet detector; with RetinaNet (Table~\ref{table:retinanetSmall}), \ourmethod-L1 has FLOPs similar to the recently proposed RegionViT-B~\citep{chen2022regionvit}, but it outperforms RegionViT-B by 2.5 and 2.0 AP$^{bb}$ in both 1$\times$ and 3$\times$ schedules with only 25\% as many parameters. In Table~\ref{table:maskrcnnSmall}, we also compare our Mask R-CNN with the recently proposed ViTDet~\citep{li2021benchmarking} that also uses a non-hierarchical ViT as the backbone network. Here we continue to use the standard 3$\times$ (36 epochs) training recipe for \ourmethod. The results show that under similar FLOPs, even if ViTDet is equipped with more parameters (111M), advanced masked-auto-encoder (MAE) pre-training~\citep{he2022masked}, a longer training schedule (100 epochs), and heavy regularizations like large-scale jittering~\citep{ghiasi2021simple}, our model can still achieve a comparable performance, which further validates the effectiveness of \ourmethod.


\begin{table}[!t]
\centering
\vspace{-4mm}
\caption{\small {Comparison between \ourmethod~and other vision transformers on ADE20K semantic segmentation task.}}
\vspace{-3mm}
\begin{subtable}[h]{0.44\textwidth}
\begin{adjustbox}{max width=\linewidth}
\LARGE
\begin{tabular}{l@{\hskip -5pt}ccc}
        \toprule
        \multicolumn{4}{c}{\LARGE UperNet} \\
        \midrule
                &  Params &  FLOPs &    \\
               Backbone &  (M) &  (G) &  mIoU  \\
        \midrule
        \multicolumn{4}{l}{Hierarchical} \\
        \midrule
        Swin-B~\citep{Swin_Liu_2021tq} &    121 &    1188 &  49.7 \\
        DAT-T~\citep{DAT} & 121 & 1212 & 50.5 \\
        DaViT-Base~\citep{DAViT} & 121 & 1175 & 49.4 \\ 
        MPViT-B~\citep{MPViT} & 105 & 1186 & 50.3 \\
        DW-B~\citep{dwvit} & 125 & 1200 & 48.7 \\
        \midrule
        \multicolumn{4}{l}{Non-hierarchical} \\
        \midrule
        Deit-B~\citep{DeiT_touvron2020} & 120 & 786 & 45.3 \\
        ViT-Adapter-B~\citep{vitadapter} & 134 & - & 48.1 \\
        \midrule
        \ourmethod-L1 &    37 & 568 &  49.1 \\
        \ourmethod-L2 &    53 & 775 &  50.2 \\
        \ourmethod-L3 &    66 & 944 &  51.7 \\
              {\ourmethod-L4} &          {107}&       {1469} &        {52.5} \\
        \bottomrule
\end{tabular}
\end{adjustbox}
\end{subtable}
\quad
\begin{subtable}[h]{0.44\textwidth}
\begin{adjustbox}{max width=\linewidth}
\LARGE
\begin{tabular}{l@{\hskip -5pt}ccc}
        \toprule
        \multicolumn{4}{c}{\LARGE SegFormer} \\
        \midrule
                &  Params &  FLOPs &    \\
               Backbone &  (M) &  (G) &  mIoU  \\
        \midrule
        \multicolumn{4}{l}{Hierarchical} \\
        \midrule
        MiT-B2~\citep{xie2021segformer} & 24 & 64 & 46.5 \\
        MiT-B4~\citep{xie2021segformer} & 61 & 96 & 50.4 \\
        Stunned-S~\citep{stunned} & 25 & 78 & 48.3 \\
        CSwin-S\citep{cswin} & 37 & 78 & 49.9 \\
        HRViT-b3\citep{gu2021hrvit} & 29 & 68 & 50.2 \\
        HILA+MiT-B1~\citep{leung2022hila} & 22 & 31 & 45 \\
        HILA+MiT-B2~\citep{leung2022hila} & 31 & 76 & 46 \\
        \midrule
        \multicolumn{4}{l}{Non-hierarchical} \\
        \midrule
        \ourmethod-L1 &    9 & 34 &  46.9 \\
        \ourmethod-L2 &    24 & 83 &  49.2 \\
        \ourmethod-L3 &    36 & 123 & 50.8\\
        {\ourmethod-L4} &          {76} &       {249} &       {51.3} \\
        \bottomrule
\end{tabular}
\end{adjustbox}
\end{subtable}
\label{table:adeCombine}
\vspace{-4mm}
\end{table}

\subsection{ADE20K Semantic Segmentation}
\textbf{Setting:} We follow previous work~\citep{Swin_Liu_2021tq} and use UperNet~\citep{xiao2018unified} as the segmentation network. We also report performance when using the recently proposed SegFormer~\citep{xie2021segformer} model. For both models, we train for 160k iterations.

\textbf{Results:} We summarise the segmentation performance of \ourmethod~and other state-of-the-art backbone networks in Table~\ref{table:adeCombine}. For UperNet, we report results with the largest available model size for the competing methods to show how far we can go in the segmentation task. Thanks to its high-resolution design, \ourmethod~outperforms all competing methods in mIoU with fewer FLOPs and fewer parameters. For example, \ourmethod-L1 only has 37M parameters but it can achieve comparable mIoU to methods with only half the number of FLOPs. This result tells us that for tasks requiring the perception of fine-grained details, scaling-up feature resolution is a better strategy than scaling up model size. \ourmethod~also excels when used with SegFormer. Specifically,~\ourmethod~achieves better mIoU than recently proposed vision transformers with similar parameter counts, including HRViT~\citep{gu2021hrvit} that was specifically designed for semantic segmentation. We attribute these promising results to \ourmethod's high-resolution design and its effective encapsulation of global information.

\vspace{-0.1in}
\subsection{Ablation Studies}

\textbf{Setting:} We conduct ablation studies using two types of local attention: the simple window attention~\citep{Swin_Liu_2021tq} and the more advanced LePE attention~\citep{cswin}, which we used in previous experiments. We use the L1 level models (Param $<$ 10 M) for all experiments. All models are pre-trained on ImageNet classification task for 300 epochs using the same setting as in Section~\ref{sec:expIn1k}. We report both ImageNet Top-1 accuracy and ADE20K SegFormer mIOU. Please refer to our appendix for more ablation experiments. 

\setlength\intextsep{0pt}
\begin{wraptable}{r}{0.45\linewidth}
    \centering
    \caption{\small {   {Ablation studies on feature resolution and \ourblock~using two different local attentions.}}}
    \label{table:stepByStep}
    \vspace{-3mm}
    \begin{adjustbox}{max width=\linewidth}
    \begin{tabular}{l|cc|cc}
        \toprule
         & FLOPs& Top-1  & FLOPs & \\
        Setting & (G) & Acc & (G) &mIoU \\
        \midrule
        ViT-D216-P16 & 1.8 & 77.4 & 16 & 42.2 \\
        ViT-D216-P8 & 8.8 & 79.2 & 113 & -  \\
        \midrule
        + Win attn & 5.8 & 78.1 & 34 & 41.7\\
        + \ourblock & 5.8 & 79.8 & 34 & 45.5 \\
        \midrule
        + LePE attn & 5.8 & 79.5 & 34 & 46.2  \\
        + \ourblock & 5.8 & 80.5 & 34 & 46.9  \\
        \bottomrule
    \end{tabular}
    \end{adjustbox}
    \vspace{2mm}
\end{wraptable}

\textbf{Building \ourmethod~step by step.} Here we show how we build \ourmethod~step by step and present the results in Table~\ref{table:stepByStep}. We start building our \ourmethod~from a low-resolution vanilla DeiT with patch sizes of 16 and embedding channels of 216 (same as \ourmethod-Tiny). It achieves 77.4 top-1 accuracy on ImageNet and 42.2 mIoU on ADE20K. Then we increase the resolution by shrinking the patch size to 8. The FLOPs of the ImageNet and ADE20K models increase by 4.4$\times$ and 7.0$\times$ respectively. ImageNet accuracy increases to 79.2 but training this model for segmentation proves to be unstable. We see that enlarging the feature resolution using global self-attention leads to the number of FLOPs exploding and makes convergence difficult. We now replace self-attention with window attention~\citep{Swin_Liu_2021tq} and the more advanced LePE attention~\citep{cswin}. For both local attention mechanisms, the FLOPs of the ImageNet and ADE20K models drop to 5.8G and 34G respectively. We then incorporate \ourblock s, and observe that the accuracy and mIoU improve for both types of local attention and FLOPs remain unchanged. These results showcase the effectiveness of using high-resolution features as well as the importance of our combination of local attention blocks and \ourblock s~to maintain a reasonable computation cost.

\setlength\intextsep{0pt}
\begin{wraptable}{r}{0.5\linewidth}
    \centering
    \caption{\small {   {Ablation study on different global information propagation methods.}}}
    \label{table:infoPropa}
    \vspace{-3mm}
    \begin{adjustbox}{max width=\linewidth}
    \LARGE
    \begin{tabular}{c|c|cc|cc}
        \toprule
         & Global & FLOPs & Top-1 & FLOPs &  \\
        Attention & Info & (G) & Acc & (G) & mIoU\\
        \midrule
        \multirow{5}{*}{Window} & None & 5.8 & 78.2 & 34 & 41.7 \\
        & Conv & 6.6 & 75.8 & 38 & 39.5 \\
        & Global attn & 6.8 & 78.8 & 62 & 44.0 \\
        & Win shift & 5.8 & 78.1 & 34 & 40.7 \\
        & \ourblock & 5.8 & 79.8 & 34 & 45.5 \\
        \midrule
        \multirow{4}{*}{LePE} & None & 5.8 & 79.5 & 34 & 46.2 \\
        & Conv & 6.6 & 77.7 & 38 & 45.5 \\
        & Global attn & 6.8 & 80.4 & 62 & 46.7\\
        & \ourblock & 5.8 & 80.5 & 34 & 46.9 \\
        \bottomrule
\end{tabular}
    \end{adjustbox}
    \vspace{2mm}
\end{wraptable}

\textbf{Global information exchange.} Here, we compare our \ourblock~with other blocks that can exchange global information between image features. The competing blocks include the global attention block, the convolution propagation block~\citep{li2022exploring}, and the shifting window-attention block~\citep{Swin_Liu_2021tq} designed for window attention. We follow ViTDet~\citep{li2022exploring} to build the convolution propagation block that stacks two 3$\times$3 convolution layers with a residual connection. We use the original version of the shifting window-attention block as in~\citet{Swin_Liu_2021tq}. The resulting models are acquired by putting competiting blocks in the same place as as our \ourblock. We report the results in Table~\ref{table:infoPropa}. We observe that simply replacing the local attention layers with convolution layers causes severe performance drops for both  types of local attention.We also observe that replacing local attention with global attention can improve performance for a very large increase in FLOPs. For window attention, we found that using the shifting window strategy slightly hurts the performance. We postulate that this is caused by a deficit of shifting window layers; half of Swin Transformer layers are shifting window layers, but we only use four here. For both types of local attention, \ourblock~achieves the best performance on ImageNet and ADE20K. These results show \ourblock's effectiveness in propagating global information. 

\setlength\intextsep{0pt}
\begin{wraptable}{r}{0.5\linewidth}
    \centering
    \caption{\small {   {Ablation study on the group tokens number combinations.}}}
    \vspace{-3mm}
    \label{table:nGroupToken}
    \begin{adjustbox}{max width=\linewidth}
    \footnotesize
    \begin{tabular}{lrr}
        \toprule
           {Combination} &    {FLOPs (G)} &    {Top-1 Acc} \\
        \midrule
           {\{16, 16, 16, 16\}} &    {5.7} &    {79.9}  \\
            {\{32, 32, 32, 32\}} &    {5.8} &    {80.3}  \\
            {\{64, 64, 64, 64\}} &     {6.0} &    {80.7}  \\
           {\{16, 32, 32, 64\}} &     {5.8} &    {80.0}  \\
           {\{64, 32, 32, 16\}} &     {5.8} &    {80.5}  \\
        \bottomrule
    \end{tabular}
    \end{adjustbox}
    \vspace{2mm}
\end{wraptable}

\textbf{Number of group tokens.}
Here we study how the different combinations of the number of groups tokens in \ourblock s affect the overall model performance. We report the results in Table~\ref{table:nGroupToken}. We find using a large number of group tokens across the whole network can give us higher accuracy on ImageNet but at additional computational cost. However, using too few group e.g.\ 16 tokens will harm performance. In \ourmethod~we choose to progressively decrease the number of group tokens from 64 to 16. This strategy gives us a good trade-off between accuracy and computational cost.

\setlength\intextsep{0pt}
\begin{wraptable}{r}{0.5\linewidth}
    \centering
    \caption{\small { {Ablation study on the propagation approach of grouped features.} }}
    \label{table:tokenPropogation}
    \vspace{-3mm}
    \begin{adjustbox}{max width=\linewidth}
    \footnotesize
    \begin{tabular}{lrr}
        \toprule
         {Method} &   {FLOPs (G)} &  {Top-1 Acc}  \\
        \midrule
         {None} &   {5.7} &  {79.8}  \\
         {SelfAttn}  &  {6.2} &  {80.7}  \\
         {MLPMixer} &  {5.8} &  {80.5} \\
        \bottomrule
    \end{tabular}
    \end{adjustbox}
    \vspace{2mm}
\end{wraptable}

\textbf{Grouped features propagation.}
In Table~\ref{table:tokenPropogation} we compare different methods for global information propagation. The results show that even when we add nothing to explicitly propagate global information the model can still achieve a good performance (79.8\% accuracy on ImageNet). The reason is that in this case the image features are still grouped and ungrouped so the global information can still be exchanged in these two operations. We also find that self-attention achieve a slightly better accuracy than MLPMixer (80.7 v.s. 80.5), but is more expensive. In \ourmethod~we use MLPMixer for propagating global information.


\section{Conclusion}
In this paper, we have presented the \ourmethodFull: a non-hierarchical  vision transformer designed for high-resolution visual recognition. The core of \ourmethod~is the \ourblock, which was proposed to efficiently exchange global information among high-resolution features. The \ourblock~first forms grouped features and thxen updates them through~\emph{Group Propagation}. Finally, these updated group features are queried back to the image features. We have shown that \ourmethod~can achieve better performance than previous work on ImageNet classification, COCO object detection and instance segmentation, and ADE20K semantic segmentation. 

\clearpage
\section{Acknowledgement}
Prof. Xiaolong Wang’s group was supported, in part, by gifts from Qualcomm and Amazon Research Award. Chenhongyi Yang was supported by a PhD studentship provided by the School of Engineering, University of Edinburgh.

\bibliography{iclr2023_conference}

\begin{thebibliography}{68}
\providecommand{\natexlab}[1]{#1}
\providecommand{\url}[1]{\texttt{#1}}
\expandafter\ifx\csname urlstyle\endcsname\relax
  \providecommand{\doi}[1]{doi: #1}\else
  \providecommand{\doi}{doi: \begingroup \urlstyle{rm}\Url}\fi

\bibitem[Carion et~al.(2020)Carion, Massa, Synnaeve, Usunier, Kirillov, and
  Zagoruyko]{detr}
Nicolas Carion, Francisco Massa, Gabriel Synnaeve, Nicolas Usunier, Alexander
  Kirillov, and Sergey Zagoruyko.
\newblock {End-to-End Object Detection with Transformers}.
\newblock In \emph{Proceedings of the European conference on computer vision},
  2020.

\bibitem[Chen et~al.(2022{\natexlab{a}})Chen, Panda, and
  Fan]{chen2022regionvit}
Chun-Fu Chen, Rameswar Panda, and Quanfu Fan.
\newblock Regionvit: Regional-to-local attention for vision transformers.
\newblock In \emph{International Conference on Learning Representations},
  2022{\natexlab{a}}.

\bibitem[Chen et~al.(2019)Chen, Wang, Pang, Cao, Xiong, Li, Sun, Feng, Liu, Xu,
  Zhang, Cheng, Zhu, Cheng, Zhao, Li, Lu, Zhu, Wu, Dai, Wang, Shi, Ouyang, Loy,
  and Lin]{mmdetection}
Kai Chen, Jiaqi Wang, Jiangmiao Pang, Yuhang Cao, Yu~Xiong, Xiaoxiao Li,
  Shuyang Sun, Wansen Feng, Ziwei Liu, Jiarui Xu, Zheng Zhang, Dazhi Cheng,
  Chenchen Zhu, Tianheng Cheng, Qijie Zhao, Buyu Li, Xin Lu, Rui Zhu, Yue Wu,
  Jifeng Dai, Jingdong Wang, Jianping Shi, Wanli Ouyang, Chen~Change Loy, and
  Dahua Lin.
\newblock {MMDetection}: Open mmlab detection toolbox and benchmark.
\newblock \emph{arXiv preprint arXiv:1906.07155}, 2019.

\bibitem[Chen et~al.(2022{\natexlab{b}})Chen, Duan, Wang, He, Lu, Dai, and
  Qiao]{vitadapter}
Zhe Chen, Yuchen Duan, Wenhai Wang, Junjun He, Tong Lu, Jifeng Dai, and
  Yu~Qiao.
\newblock Vision transformer adapter for dense predictions.
\newblock \emph{arXiv preprint arXiv:2205.08534}, 2022{\natexlab{b}}.

\bibitem[Chen et~al.(2021)Chen, Xie, Niu, Liu, Wei, and
  Tian]{Visformer_Chen_2021ve}
Zhengsu Chen, Lingxi Xie, Jianwei Niu, Xuefeng Liu, Longhui Wei, and Qi~Tian.
\newblock Visformer: The vision-friendly transformer.
\newblock In \emph{Proceedings of the IEEE/CVF International Conference on
  Computer Vision}, pp.\  589--598, October 2021.

\bibitem[Cheng et~al.(2020)Cheng, Xiao, Wang, Shi, Huang, and
  Zhang]{HigherHRNet_Cheng_2020_CVPR}
Bowen Cheng, Bin Xiao, Jingdong Wang, Honghui Shi, Thomas~S. Huang, and Lei
  Zhang.
\newblock Higherhrnet: Scale-aware representation learning for bottom-up human
  pose estimation.
\newblock In \emph{IEEE/CVF Conference on Computer Vision and Pattern
  Recognition}, June 2020.

\bibitem[Cheng et~al.(2022)Cheng, Misra, Schwing, Kirillov, and
  Girdhar]{cheng2021mask2former}
Bowen Cheng, Ishan Misra, Alexander~G. Schwing, Alexander Kirillov, and Rohit
  Girdhar.
\newblock Masked-attention mask transformer for universal image segmentation.
\newblock In \emph{Proceedings of the IEEE/CVF Conference on Computer Vision
  and Pattern Recognition}, June 2022.

\bibitem[Contributors(2020{\natexlab{a}})]{2020mmclassification}
MMClassification Contributors.
\newblock Openmmlab's image classification toolbox and benchmark.
\newblock \url{https://github.com/open-mmlab/mmclassification},
  2020{\natexlab{a}}.

\bibitem[Contributors(2020{\natexlab{b}})]{mmseg2020}
MMSegmentation Contributors.
\newblock {MMSegmentation}: Openmmlab semantic segmentation toolbox and
  benchmark.
\newblock \url{https://github.com/open-mmlab/mmsegmentation},
  2020{\natexlab{b}}.

\bibitem[Cubuk et~al.(2020)Cubuk, Zoph, Shlens, and Le]{cubuk2020randaugment}
Ekin~D Cubuk, Barret Zoph, Jonathon Shlens, and Quoc~V Le.
\newblock Randaugment: Practical automated data augmentation with a reduced
  search space.
\newblock In \emph{Proceedings of the IEEE/CVF conference on computer vision
  and pattern recognition workshops}, pp.\  702--703, 2020.

\bibitem[Ding et~al.(2022)Ding, Xiao, Codella, Luo, Wang, and Yuan]{DAViT}
Mingyu Ding, Bin Xiao, Noel Codella, Ping Luo, Jingdong Wang, and Lu~Yuan.
\newblock Davit: Dual attention vision transformers.
\newblock In \emph{Proceedings of the European conference on computer vision},
  2022.

\bibitem[Dong et~al.(2022)Dong, Bao, Chen, Zhang, Yu, Yuan, Chen, and
  Guo]{cswin}
Xiaoyi Dong, Jianmin Bao, Dongdong Chen, Weiming Zhang, Nenghai Yu, Lu~Yuan,
  Dong Chen, and Baining Guo.
\newblock Cswin transformer: A general vision transformer backbone with
  cross-shaped windows.
\newblock In \emph{Proceedings of the IEEE/CVF Conference on Computer Vision
  and Pattern Recognition}, pp.\  12124--12134, June 2022.

\bibitem[Dosovitskiy et~al.(2021)Dosovitskiy, Beyer, Kolesnikov, Weissenborn,
  Zhai, Unterthiner, Dehghani, Minderer, Heigold, Gelly, Uszkoreit, and
  Houlsby]{ViT_dosovitskiy2021an}
Alexey Dosovitskiy, Lucas Beyer, Alexander Kolesnikov, Dirk Weissenborn,
  Xiaohua Zhai, Thomas Unterthiner, Mostafa Dehghani, Matthias Minderer, Georg
  Heigold, Sylvain Gelly, Jakob Uszkoreit, and Neil Houlsby.
\newblock An image is worth 16x16 words: Transformers for image recognition at
  scale.
\newblock In \emph{International Conference on Learning Representations}, 2021.
\newblock URL \url{https://openreview.net/forum?id=YicbFdNTTy}.

\bibitem[d’Ascoli et~al.(2021)d’Ascoli, Touvron, Leavitt, Morcos, Biroli,
  and Sagun]{ConViTdAscoli_2021vz}
St{\'e}phane d’Ascoli, Hugo Touvron, Matthew~L Leavitt, Ari~S Morcos, Giulio
  Biroli, and Levent Sagun.
\newblock Convit: Improving vision transformers with soft convolutional
  inductive biases.
\newblock In \emph{International Conference on Machine Learning}. PMLR, 2021.

\bibitem[Elsayed et~al.(2022)Elsayed, Mahendran, van Steenkiste, Greff, Mozer,
  and Kipf]{elsayed2022savi++}
Gamaleldin~F. Elsayed, Aravindh Mahendran, Sjoerd van Steenkiste, Klaus Greff,
  Michael~C. Mozer, and Thomas Kipf.
\newblock {SAVi++}: Towards end-to-end object-centric learning from real-world
  videos.
\newblock In \emph{Advances in Neural Information Processing Systems}, 2022.

\bibitem[Fan et~al.(2021)Fan, Xiong, Mangalam, Li, Yan, Malik, and
  Feichtenhofer]{fan2021multiscale}
Haoqi Fan, Bo~Xiong, Karttikeya Mangalam, Yanghao Li, Zhicheng Yan, Jitendra
  Malik, and Christoph Feichtenhofer.
\newblock Multiscale vision transformers.
\newblock In \emph{Proceedings of the IEEE/CVF International Conference on
  Computer Vision}, pp.\  6824--6835, 2021.

\bibitem[Ghiasi et~al.(2021)Ghiasi, Cui, Srinivas, Qian, Lin, Cubuk, Le, and
  Zoph]{ghiasi2021simple}
Golnaz Ghiasi, Yin Cui, Aravind Srinivas, Rui Qian, Tsung-Yi Lin, Ekin~D Cubuk,
  Quoc~V Le, and Barret Zoph.
\newblock Simple copy-paste is a strong data augmentation method for instance
  segmentation.
\newblock In \emph{Proceedings of the IEEE/CVF Conference on Computer Vision
  and Pattern Recognition}, pp.\  2918--2928, 2021.

\bibitem[Graham et~al.(2021)Graham, El-Nouby, Touvron, Stock, Joulin, J\'egou,
  and Douze]{LeViT_BenGraham_2021vh}
Benjamin Graham, Alaaeldin El-Nouby, Hugo Touvron, Pierre Stock, Armand Joulin,
  Herv\'e J\'egou, and Matthijs Douze.
\newblock Levit: A vision transformer in convnet's clothing for faster
  inference.
\newblock In \emph{Proceedings of the IEEE/CVF International Conference on
  Computer Vision}, 2021.

\bibitem[Gu et~al.(2022)Gu, Kwon, Wang, Ye, Li, Chen, Lai, Chandra, and
  Pan]{gu2021hrvit}
Jiaqi Gu, Hyoukjun Kwon, Dilin Wang, Wei Ye, Meng Li, Yu-Hsin Chen, Liangzhen
  Lai, Vikas Chandra, and David~Z. Pan.
\newblock Multi-scale high-resolution vision transformer for semantic
  segmentation.
\newblock In \emph{Proceedings of the IEEE/CVF Conference on Computer Vision
  and Pattern Recognition}, 2022.

\bibitem[Guo et~al.(2022)Guo, Han, Wu, Tang, Chen, Wang, and Xu]{CMT}
Jianyuan Guo, Kai Han, Han Wu, Yehui Tang, Xinghao Chen, Yunhe Wang, and Chang
  Xu.
\newblock Cmt: Convolutional neural networks meet vision transformers.
\newblock In \emph{Proceedings of the IEEE/CVF Conference on Computer Vision
  and Pattern Recognition}, pp.\  12175--12185, June 2022.

\bibitem[Han et~al.(2021)Han, Xiao, Wu, Guo, Xu, and Wang]{han2021transformer}
Kai Han, An~Xiao, Enhua Wu, Jianyuan Guo, Chunjing Xu, and Yunhe Wang.
\newblock Transformer in transformer.
\newblock \emph{Advances in Neural Information Processing Systems},
  34:\penalty0 15908--15919, 2021.

\bibitem[Hatamizadeh et~al.(2022)Hatamizadeh, Yin, Kautz, and Molchanov]{gcvit}
Ali Hatamizadeh, Hongxu Yin, Jan Kautz, and Pavlo Molchanov.
\newblock Global context vision transformers.
\newblock \emph{arXiv preprint arXiv:2206.09959}, 2022.

\bibitem[He et~al.(2016)He, Zhang, Ren, and Sun]{ResNet_He_2016_CVPR}
Kaiming He, Xiangyu Zhang, Shaoqing Ren, and Jian Sun.
\newblock {Deep Residual Learning for Image Recognition}.
\newblock In \emph{The IEEE Conference on Computer Vision and Pattern
  Recognition}, June 2016.

\bibitem[He et~al.(2017)He, Gkioxari, Dollar, and
  Girshick]{MaskRCNN_He_2017_ICCV}
Kaiming He, Georgia Gkioxari, Piotr Dollar, and Ross Girshick.
\newblock Mask r-cnn.
\newblock In \emph{Proceedings of the IEEE International Conference on Computer
  Vision}, Oct 2017.

\bibitem[He et~al.(2022)He, Chen, Xie, Li, Doll{\'a}r, and
  Girshick]{he2022masked}
Kaiming He, Xinlei Chen, Saining Xie, Yanghao Li, Piotr Doll{\'a}r, and Ross
  Girshick.
\newblock Masked autoencoders are scalable vision learners.
\newblock In \emph{Proceedings of the IEEE/CVF Conference on Computer Vision
  and Pattern Recognition}, pp.\  16000--16009, 2022.

\bibitem[Huang et~al.(2019)Huang, Wang, Huang, Huang, Wei, and
  Liu]{huang2019ccnet}
Zilong Huang, Xinggang Wang, Lichao Huang, Chang Huang, Yunchao Wei, and Wenyu
  Liu.
\newblock Ccnet: Criss-cross attention for semantic segmentation.
\newblock In \emph{Proceedings of the IEEE/CVF international conference on
  computer vision}, pp.\  603--612, 2019.

\bibitem[Hudson \& Zitnick(2021)Hudson and Zitnick]{hudson2021generative}
Drew~A Hudson and Larry Zitnick.
\newblock Generative adversarial transformers.
\newblock In \emph{International conference on machine learning}, pp.\
  4487--4499. PMLR, 2021.

\bibitem[Jaegle et~al.(2022)Jaegle, Borgeaud, Alayrac, Doersch, Ionescu, Ding,
  Koppula, Zoran, Brock, Shelhamer, Henaff, Botvinick, Zisserman, Vinyals, and
  Carreira]{jaegle2021perceiver}
Andrew Jaegle, Sebastian Borgeaud, Jean-Baptiste Alayrac, Carl Doersch, Catalin
  Ionescu, David Ding, Skanda Koppula, Daniel Zoran, Andrew Brock, Evan
  Shelhamer, Olivier~J Henaff, Matthew Botvinick, Andrew Zisserman, Oriol
  Vinyals, and Joao Carreira.
\newblock Perceiver {IO}: A general architecture for structured inputs \&
  outputs.
\newblock In \emph{International Conference on Learning Representations}, 2022.

\bibitem[Kipf et~al.(2022)Kipf, Elsayed, Mahendran, Stone, Sabour, Heigold,
  Jonschkowski, Dosovitskiy, and Greff]{kipf2021conditional}
Thomas Kipf, Gamaleldin~F. Elsayed, Aravindh Mahendran, Austin Stone, Sara
  Sabour, Georg Heigold, Rico Jonschkowski, Alexey Dosovitskiy, and Klaus
  Greff.
\newblock {Conditional Object-Centric Learning from Video}.
\newblock In \emph{International Conference on Learning Representations
  (ICLR)}, 2022.

\bibitem[Kipf \& Welling(2017)Kipf and Welling]{kipf2016semi}
Thomas~N. Kipf and Max Welling.
\newblock Semi-supervised classification with graph convolutional networks.
\newblock In \emph{International Conference on Learning Representations}, 2017.

\bibitem[Lee et~al.(2022)Lee, Kim, Willette, and Hwang]{MPViT}
Youngwan Lee, Jonghee Kim, Jeffrey Willette, and Sung~Ju Hwang.
\newblock Mpvit: Multi-path vision transformer for dense prediction.
\newblock In \emph{Proceedings of the IEEE/CVF Conference on Computer Vision
  and Pattern Recognition}, pp.\  7287--7296, June 2022.

\bibitem[Leung et~al.(2022)Leung, Gao, Zeng, and Fidler]{leung2022hila}
Gary Leung, Jun Gao, Xiaohui Zeng, and Sanja Fidler.
\newblock Hila: Improving semantic segmentation in transformers using
  hierarchical inter-level attention.
\newblock \emph{arXiv:2207.02126}, 2022.

\bibitem[Li et~al.(2021{\natexlab{a}})Li, Xie, Chen, Dollar, He, and
  Girshick]{li2021benchmarking}
Yanghao Li, Saining Xie, Xinlei Chen, Piotr Dollar, Kaiming He, and Ross
  Girshick.
\newblock Benchmarking detection transfer learning with vision transformers.
\newblock \emph{arXiv preprint arXiv:2111.11429}, 2021{\natexlab{a}}.

\bibitem[Li et~al.(2022{\natexlab{a}})Li, Mao, Girshick, and
  He]{li2022exploring}
Yanghao Li, Hanzi Mao, Ross Girshick, and Kaiming He.
\newblock Exploring plain vision transformer backbones for object detection.
\newblock In \emph{Proceedings of the IEEE conference on computer vision and
  pattern recognition}, 2022{\natexlab{a}}.

\bibitem[Li et~al.(2022{\natexlab{b}})Li, Wu, Fan, Mangalam, Xiong, Malik, and
  Feichtenhofer]{mvitv2}
Yanghao Li, Chao-Yuan Wu, Haoqi Fan, Karttikeya Mangalam, Bo~Xiong, Jitendra
  Malik, and Christoph Feichtenhofer.
\newblock Mvitv2: Improved multiscale vision transformers for classification
  and detection.
\newblock In \emph{Proceedings of the IEEE/CVF Conference on Computer Vision
  and Pattern Recognition}, pp.\  4804--4814, June 2022{\natexlab{b}}.

\bibitem[Li et~al.(2021{\natexlab{b}})Li, Zhang, Cao, Timofte, and
  Van~Gool]{LocalViT_Li_2021ww}
Yawei Li, Kai Zhang, Jiezhang Cao, Radu Timofte, and Luc Van~Gool.
\newblock {LocalViT: Bringing Locality to Vision Transformers}.
\newblock \emph{arXiv.org}, April 2021{\natexlab{b}}.

\bibitem[Li \& Gupta(2018)Li and Gupta]{li2018beyond}
Yin Li and Abhinav Gupta.
\newblock Beyond grids: Learning graph representations for visual recognition.
\newblock \emph{Advances in Neural Information Processing Systems}, 31, 2018.

\bibitem[Liu et~al.(2021)Liu, Lin, Cao, Hu, Wei, Zhang, Lin, and
  Guo]{Swin_Liu_2021tq}
Ze~Liu, Yutong Lin, Yue Cao, Han Hu, Yixuan Wei, Zheng Zhang, Stephen Lin, and
  Baining Guo.
\newblock {Swin Transformer: Hierarchical Vision Transformer using Shifted
  Windows}.
\newblock In \emph{Proceedings of the IEEE/CVF International Conference on
  Computer Vision}, 2021.

\bibitem[Locatello et~al.(2020)Locatello, Weissenborn, Unterthiner, Mahendran,
  Heigold, Uszkoreit, Dosovitskiy, and Kipf]{locatello2020object}
Francesco Locatello, Dirk Weissenborn, Thomas Unterthiner, Aravindh Mahendran,
  Georg Heigold, Jakob Uszkoreit, Alexey Dosovitskiy, and Thomas Kipf.
\newblock Object-centric learning with slot attention.
\newblock \emph{Advances in Neural Information Processing Systems},
  33:\penalty0 11525--11538, 2020.

\bibitem[Qi et~al.(2021)Qi, Wang, Pathak, Ma, and Malik]{qi2020learning}
Haozhi Qi, Xiaolong Wang, Deepak Pathak, Yi~Ma, and Jitendra Malik.
\newblock Learning long-term visual dynamics with region proposal interaction
  networks.
\newblock In \emph{International Conference on Learning Representations}, 2021.

\bibitem[Ren et~al.(2022{\natexlab{a}})Ren, Li, Wang, Xiao, Du, Liang, and
  Chang]{dwvit}
Pengzhen Ren, Changlin Li, Guangrun Wang, Yun Xiao, Qing Du, Xiaodan Liang, and
  Xiaojun Chang.
\newblock Beyond fixation: Dynamic window visual transformer.
\newblock In \emph{Proceedings of the IEEE/CVF Conference on Computer Vision
  and Pattern Recognition}, pp.\  11987--11997, June 2022{\natexlab{a}}.

\bibitem[Ren et~al.(2022{\natexlab{b}})Ren, Zhou, He, Feng, and Wang]{stunned}
Sucheng Ren, Daquan Zhou, Shengfeng He, Jiashi Feng, and Xinchao Wang.
\newblock Shunted self-attention via multi-scale token aggregation.
\newblock In \emph{Proceedings of the IEEE/CVF Conference on Computer Vision
  and Pattern Recognition}, pp.\  10853--10862, June 2022{\natexlab{b}}.

\bibitem[Sun et~al.(2019)Sun, Xiao, Liu, and Wang]{sun2019deep}
Ke~Sun, Bin Xiao, Dong Liu, and Jingdong Wang.
\newblock Deep high-resolution representation learning for human pose
  estimation.
\newblock In \emph{Proceedings of the IEEE/CVF conference on computer vision
  and pattern recognition}, pp.\  5693--5703, 2019.

\bibitem[Tolstikhin et~al.(2021)Tolstikhin, Houlsby, Kolesnikov, Beyer, Zhai,
  Unterthiner, Yung, Steiner, Keysers, Uszkoreit, et~al.]{tolstikhin2021mlp}
Ilya~O Tolstikhin, Neil Houlsby, Alexander Kolesnikov, Lucas Beyer, Xiaohua
  Zhai, Thomas Unterthiner, Jessica Yung, Andreas Steiner, Daniel Keysers,
  Jakob Uszkoreit, et~al.
\newblock Mlp-mixer: An all-mlp architecture for vision.
\newblock \emph{Advances in Neural Information Processing Systems},
  34:\penalty0 24261--24272, 2021.

\bibitem[Touvron et~al.(2021{\natexlab{a}})Touvron, Cord, Douze, Massa,
  Sablayrolles, and J\'egou]{DeiT_touvron2020}
Hugo Touvron, Matthieu Cord, Matthijs Douze, Francisco Massa, Alexandre
  Sablayrolles, and Herv\'e J\'egou.
\newblock Training data-efficient image transformers \& distillation through
  attention.
\newblock In \emph{Proceedings of the 38th International Conference on Machine
  Learning}, 2021{\natexlab{a}}.

\bibitem[Touvron et~al.(2021{\natexlab{b}})Touvron, Cord, Sablayrolles,
  Synnaeve, and J\'egou]{CaiT_Touvron:2021tp}
Hugo Touvron, Matthieu Cord, Alexandre Sablayrolles, Gabriel Synnaeve, and
  Herv\'e J\'egou.
\newblock Going deeper with image transformers.
\newblock In \emph{Proceedings of the IEEE/CVF International Conference on
  Computer Vision}, 2021{\natexlab{b}}.

\bibitem[Vaswani et~al.(2017)Vaswani, Shazeer, Parmar, Uszkoreit, Jones, Gomez,
  Kaiser, and Polosukhin]{Transformer_NIPS2017_Vaswani}
Ashish Vaswani, Noam Shazeer, Niki Parmar, Jakob Uszkoreit, Llion Jones,
  Aidan~N Gomez, ukasz Kaiser, and Illia Polosukhin.
\newblock {Attention is All you Need}.
\newblock In I~Guyon, U~V Luxburg, S~Bengio, H~Wallach, R~Fergus,
  S~Vishwanathan, and R~Garnett (eds.), \emph{Advances in Neural Information
  Processing Systems}. Curran Associates, Inc., 2017.

\bibitem[Wang et~al.(2022{\natexlab{a}})Wang, Zhao, Tang, Luo, and
  Zeng]{shiftvit}
Guangting Wang, Yucheng Zhao, Chuanxin Tang, Chong Luo, and Wenjun Zeng.
\newblock When shift operation meets vision transformer: An extremely simple
  alternative to attention mechanism.
\newblock In \emph{AAAI Conference on Artificial Intelligence},
  2022{\natexlab{a}}.

\bibitem[Wang et~al.(2020)Wang, Sun, Cheng, Jiang, Deng, Zhao, Liu, Mu, Tan,
  Wang, et~al.]{wang2020deep}
Jingdong Wang, Ke~Sun, Tianheng Cheng, Borui Jiang, Chaorui Deng, Yang Zhao,
  Dong Liu, Yadong Mu, Mingkui Tan, Xinggang Wang, et~al.
\newblock Deep high-resolution representation learning for visual recognition.
\newblock \emph{IEEE transactions on pattern analysis and machine
  intelligence}, 43\penalty0 (10):\penalty0 3349--3364, 2020.

\bibitem[Wang et~al.(2021)Wang, Xie, Li, Fan, Song, Liang, Lu, Luo, and
  Shao]{PVT_wang2021}
Wenhai Wang, Enze Xie, Xiang Li, Deng-Ping Fan, Kaitao Song, Ding Liang, Tong
  Lu, Ping Luo, and Ling Shao.
\newblock Pyramid vision transformer: A versatile backbone for dense prediction
  without convolutions.
\newblock In \emph{Proceedings of the IEEE/CVF International Conference on
  Computer Vision}, 2021.

\bibitem[Wang et~al.(2022{\natexlab{b}})Wang, Xie, Li, Fan, Song, Liang, Lu,
  Luo, and Shao]{pvtv2}
Wenhai Wang, Enze Xie, Xiang Li, Deng-Ping Fan, Kaitao Song, Ding Liang, Tong
  Lu, Ping Luo, and Ling Shao.
\newblock Pvtv2: Improved baselines with pyramid vision transformer.
\newblock \emph{Computational Visual Media}, 2022{\natexlab{b}}.

\bibitem[Wang \& Gupta(2018)Wang and Gupta]{wang2018videos}
Xiaolong Wang and Abhinav Gupta.
\newblock Videos as space-time region graphs.
\newblock In \emph{Proceedings of the European conference on computer vision},
  pp.\  399--417, 2018.

\bibitem[Watters et~al.(2017)Watters, Zoran, Weber, Battaglia, Pascanu, and
  Tacchetti]{watters2017visual}
Nicholas Watters, Daniel Zoran, Theophane Weber, Peter Battaglia, Razvan
  Pascanu, and Andrea Tacchetti.
\newblock Visual interaction networks: Learning a physics simulator from video.
\newblock \emph{Advances in neural information processing systems}, 30, 2017.

\bibitem[Wu et~al.(2021)Wu, Xiao, Codella, Liu, Dai, Yuan, and
  Zhang]{CvT_Wu_2021tw}
Haiping Wu, Bin Xiao, Noel Codella, Mengchen Liu, Xiyang Dai, Lu~Yuan, and Lei
  Zhang.
\newblock Cvt: Introducing convolutions to vision transformers.
\newblock In \emph{Proceedings of the IEEE/CVF International Conference on
  Computer Vision}, 2021.

\bibitem[Xia et~al.(2022)Xia, Pan, Song, Li, and Huang]{DAT}
Zhuofan Xia, Xuran Pan, Shiji Song, Li~Erran Li, and Gao Huang.
\newblock Vision transformer with deformable attention.
\newblock In \emph{Proceedings of the IEEE/CVF Conference on Computer Vision
  and Pattern Recognition}, pp.\  4794--4803, June 2022.

\bibitem[Xiao et~al.(2018)Xiao, Liu, Zhou, Jiang, and Sun]{xiao2018unified}
Tete Xiao, Yingcheng Liu, Bolei Zhou, Yuning Jiang, and Jian Sun.
\newblock Unified perceptual parsing for scene understanding.
\newblock In \emph{Proceedings of the European conference on computer vision},
  pp.\  418--434, 2018.

\bibitem[Xie et~al.(2021)Xie, Wang, Yu, Anandkumar, Alvarez, and
  Luo]{xie2021segformer}
Enze Xie, Wenhai Wang, Zhiding Yu, Anima Anandkumar, Jose~M Alvarez, and Ping
  Luo.
\newblock Segformer: Simple and efficient design for semantic segmentation with
  transformers.
\newblock \emph{Advances in Neural Information Processing Systems},
  34:\penalty0 12077--12090, 2021.

\bibitem[Xie et~al.(2017)Xie, Girshick, Doll{\'a}r, Tu, and
  He]{ResNeXt_Xie_2017_CVPR}
Saining Xie, Ross Girshick, Piotr Doll{\'a}r, Zhuowen Tu, and Kaiming He.
\newblock {Aggregated Residual Transformations for Deep Neural Networks}.
\newblock In \emph{The IEEE Conference on Computer Vision and Pattern
  Recognition}, July 2017.

\bibitem[Xu et~al.(2022)Xu, De~Mello, Liu, Byeon, Breuel, Kautz, and
  Wang]{xu2022groupvit}
Jiarui Xu, Shalini De~Mello, Sifei Liu, Wonmin Byeon, Thomas Breuel, Jan Kautz,
  and Xiaolong Wang.
\newblock Groupvit: Semantic segmentation emerges from text supervision.
\newblock In \emph{Proceedings of the IEEE/CVF Conference on Computer Vision
  and Pattern Recognition}, 2022.

\bibitem[Xu et~al.(2021{\natexlab{a}})Xu, Xu, Chang, and Tu]{xu2021co}
Weijian Xu, Yifan Xu, Tyler Chang, and Zhuowen Tu.
\newblock Co-scale conv-attentional image transformers.
\newblock In \emph{Proceedings of the IEEE/CVF International Conference on
  Computer Vision}, pp.\  9981--9990, 2021{\natexlab{a}}.

\bibitem[Xu et~al.(2021{\natexlab{b}})Xu, Zhang, Zhang, and Tao]{xu2021vitae}
Yufei Xu, Qiming Zhang, Jing Zhang, and Dacheng Tao.
\newblock Vitae: Vision transformer advanced by exploring intrinsic inductive
  bias.
\newblock \emph{Advances in Neural Information Processing Systems},
  34:\penalty0 28522--28535, 2021{\natexlab{b}}.

\bibitem[Yang et~al.(2022)Yang, Huang, and Wang]{Yang_2022_CVPR}
Chenhongyi Yang, Zehao Huang, and Naiyan Wang.
\newblock Querydet: Cascaded sparse query for accelerating high-resolution
  small object detection.
\newblock In \emph{Proceedings of the IEEE/CVF Conference on Computer Vision
  and Pattern Recognition}, pp.\  13668--13677, June 2022.

\bibitem[Yang et~al.(2021)Yang, Li, Zhang, Dai, Xiao, Yuan, and Gao]{focal}
Jianwei Yang, Chunyuan Li, Pengchuan Zhang, Xiyang Dai, Bin Xiao, Lu~Yuan, and
  Jianfeng Gao.
\newblock Focal self-attention for local-global interactions in vision
  transformers.
\newblock In \emph{NeurIPS}, 2021.

\bibitem[Yuan et~al.(2021)Yuan, Fu, Huang, Lin, Zhang, Chen, and
  Wang]{yuan2021hrformer}
Yuhui Yuan, Rao Fu, Lang Huang, Weihong Lin, Chao Zhang, Xilin Chen, and
  Jingdong Wang.
\newblock Hrformer: High-resolution vision transformer for dense predict.
\newblock \emph{Advances in Neural Information Processing Systems},
  34:\penalty0 7281--7293, 2021.

\bibitem[Yun et~al.(2019)Yun, Han, Oh, Chun, Choe, and Yoo]{yun2019cutmix}
Sangdoo Yun, Dongyoon Han, Seong~Joon Oh, Sanghyuk Chun, Junsuk Choe, and
  Youngjoon Yoo.
\newblock Cutmix: Regularization strategy to train strong classifiers with
  localizable features.
\newblock In \emph{Proceedings of the IEEE/CVF international conference on
  computer vision}, pp.\  6023--6032, 2019.

\bibitem[Zhang et~al.(2017)Zhang, Cisse, Dauphin, and
  Lopez-Paz]{zhang2017mixup}
Hongyi Zhang, Moustapha Cisse, Yann~N Dauphin, and David Lopez-Paz.
\newblock mixup: Beyond empirical risk minimization.
\newblock \emph{arXiv preprint arXiv:1710.09412}, 2017.

\bibitem[Zhang et~al.(2022)Zhang, Zhang, Zhao, Chen, Arik, and
  Pfister]{zhang2022nested}
Zizhao Zhang, Han Zhang, Long Zhao, Ting Chen, Sercan~{\"O} Arik, and Tomas
  Pfister.
\newblock Nested hierarchical transformer: Towards accurate, data-efficient and
  interpretable visual understanding.
\newblock In \emph{Proceedings of the AAAI Conference on Artificial
  Intelligence}, volume~36, pp.\  3417--3425, 2022.

\bibitem[Zhong et~al.(2020)Zhong, Zheng, Kang, Li, and Yang]{zhong2020random}
Zhun Zhong, Liang Zheng, Guoliang Kang, Shaozi Li, and Yi~Yang.
\newblock Random erasing data augmentation.
\newblock In \emph{Proceedings of the AAAI conference on artificial
  intelligence}, 2020.

\end{thebibliography}
\bibliographystyle{iclr2023_conference}

\clearpage
\appendix
\vspace{-0.1in}
\section{Further Ablation studies}
\vspace{-0.1in}

\begin{figure*}[!t]
    \centering
    \includegraphics[width=\textwidth]{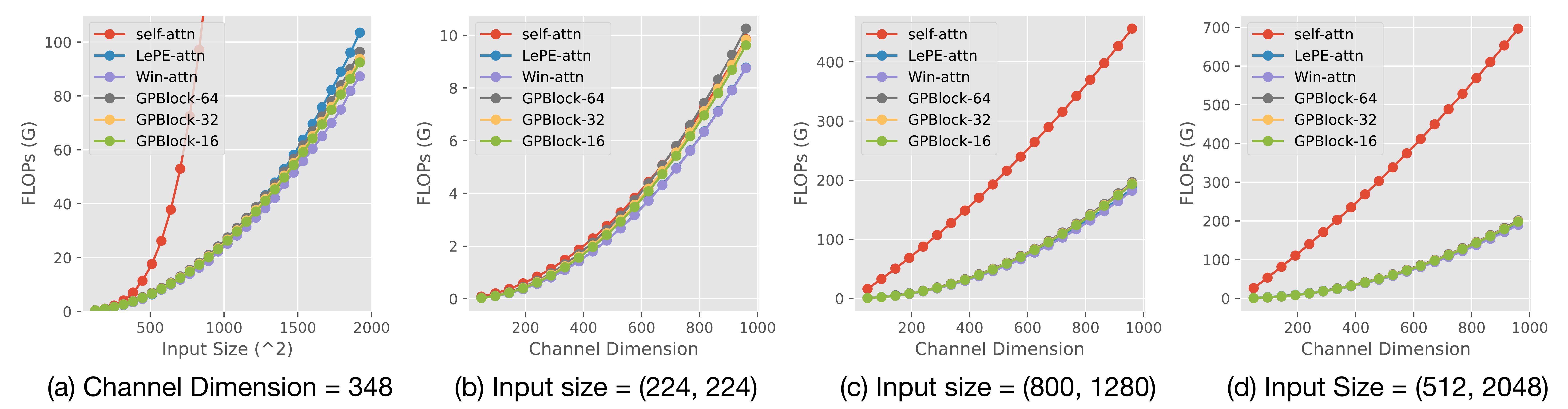}
    \vspace{-0.5cm}
    \captionsetup{font={small}}
\caption{Layer-wise FLOPs comparison between different attention blocks on high-resolution features. ``GPBlock-$N$'' denotes \ourblock~with $N$ group tokens. (a) FLOPs v.s. Input size when using feature channel dimension of 348 (same as GPViT-s1); (b-d): FLOPs v.s. feature channel dimension on three typical input sizes using by ImageNet classification, COCO object detection and instance segmentation, and ADE20K semantic segmentation. Our \ourblock~can effectively gather and propagate global information like a self-attention block while having similar or even fewer FLOPs with local attention blocks when the model or input scale-up.}
\label{fig:flops}
\end{figure*}

\begin{table}[tb]
\centering
\caption{\small Inference speed comparison on three typical input sizes: 224$\times$224 for ImageNet-1K classification, 800$\times$1280 for COCO object detection and instance segmentation, and $512\times2048$ for ADE20K semantic segmentation. The results are evaluated on NVIDIA 2080Ti GPUs. (`OOM' denotes out of GPU memory.)}
\label{table:speed}
\vspace{-3mm}
\begin{adjustbox}{max width=0.6\linewidth}
\tablestyle{2px}{1.0}
{\begin{tabular}{l@{\hskip 10pt}|c|ccc}
\toprule
      &       Param      &     \multicolumn{3}{|c}{Inference Time (ms)} \\
Model &   (M) & 224$\times$224 & 800$\times$1280  & 512$\times$2048 \\
\midrule
Low-resolution Baseline & & & &\\
ViT-D216-P16 & 7.4 & 0.9  & 155 & 150 \\
ViT-D348-P16 & 18.5 & 1.4 & 167 & 162 \\
ViT-D432-P16 & 28.5 & 1.8 & 177 & 173 \\
ViT-D624-P16 & 57.9 & 2.9 & 206 & 203 \\
\midrule
High-resolution Baseline & & & & \\
ViT-D216-P8 & 7.4 & 7.5 & OOM & OOM \\
ViT-D348-P8 & 18.5 & 9.5 & OOM & OOM \\
ViT-D432-P8 & 28.5 & 11.3 & OOM & OOM \\
ViT-D624-P8 & 57.9 & 15.9 & OOM & OOM \\
\midrule
\ourmethod-L1 & 9.3 &  2.7 & 83 & 87 \\
\ourmethod-L2 & 23.8 &  5.1 & 132 & 137 \\
\ourmethod-L3 & 36.2 &   7.0 & 174 & 182 \\
\ourmethod-L4 & 75.4 & 11.1 & 281 & 290 \\
\bottomrule
\end{tabular}}
\end{adjustbox}
\vspace{-5mm}
\end{table}

\subsection{Study on running efficiency}
\vspace{-0.1in}
  {In Table~\ref{table:speed}, we compare the inference speed of \ourmethod~with ViT baselines. Specifically, for each variant of our \ourmethod, we compare it to ViT models with patch size 16 (low-resolution) and patch size 8 (high-resolution) while keeping the channel dimensions the same. We report inference time using three different input sizes, which correspond to the three typical input sizes used by ImageNet-1k classification, COCO object detection and instance segmentation, and ADE20K semantic segmentation. We draw three observations from the results: 
\begin{itemize}
    \item When using small-sized inputs, \ourmethod~runs slower than the low-resolution ViT. Despite the efficient design of our~\ourblock~and the use of local attention, the high-resolution design still incurs a significant cost for forward passes, slowing down inference speed.
    \item When the models are applied to downstream tasks where they take larger-sized inputs, all but the largest \ourmethod~models are faster than their low-resolution ViT counterparts. For example, when the model channel number is 216, \ourmethod~takes 83 ms to process an 800$\times$1280 sized image, while ViT-D216-P16 takes 155 ms. In this case, the self-attention operations with quadratic complexity severely slow down the speed of the ViT even with low resolution features. On the other hand, the computations in \ourblock~and local attentions grow much less than self-attention when the input scales up.
    \item \ourmethod~is faster than the high-resolution ViT baselines when using small inputs. In addition, high-resolution ViTs are not even able to process large-sized inputs: we got \emph{Out of Memory} errors when using a NVIDIA 2080Ti GPU with 11 GB memory. This highlights our technical contribution of efficiently processing high-resolution features with~\ourmethod.
\end{itemize}
    }

  {We further study how the computation cost for high-resolution features changes when the model size and input scale up by examining FLOP counts. The results are shown in Figure~\ref{fig:flops} where we compare \ourblock~with different group numbers to self-attention and local-attention operations: Self-attention and \ourblock~can both exchange global information between image features, but the computational cost of \ourblock~grows much slower than self-attention. Local attention operations have a similar level of efficiency to \ourblock, but are unable to exchange global information because of their limited receptive field.}

\vspace{-0.1in}
\section{Implementation Details}
\vspace{-0.1in}
\subsection{Model Details of \ourmethod}
\label{sec:modelDetails}
The model details of different \ourmethod~variants are presented in Table~\ref{table:modelDetials}. Different \ourmethod~variants are main difference by their model width (channels) and share similar hyper-parameters in other architecture designs. 

\begin{table}[!ht]
\centering
\caption{\small \ourmethod~model details for different variants.}
\label{table:modelDetials}
\vspace{-2mm}
\begin{adjustbox}{max width=0.9\linewidth}
\tablestyle{2pt}{0.95}
\begin{tabular}{l|c|c|c|c} 
\toprule
Variants & \ourmethod-L1 & \ourmethod-L2 & \ourmethod-L3 &   {\ourmethod-L4}\\
\midrule
Patch Size & 8 & 8 & 8 &   {8}\\
Channel Dimension & 216 & 348 & 432 &   {624} \\
Number of Transformer Layers & 12 & 12 & 12 &   {12}\\ 
LePE Strip Size & 2 & 2 & 2 &   {2} \\ 
Attention Heads & 12 & 12 & 12 &   {12}\\
FFN Expansion & 4 & 4 & 4 &   {4} \\
\ourblock~Positions & \{1, 4, 7, 10\} & \{1, 4, 7, 10\} & \{1, 4, 7, 10\} &   {\{1, 4, 7, 10\}} \\
\ourblock~Group Numbers & \{64, 32, 32, 16\} & \{64, 32, 32, 16\} & \{64, 32, 32, 16\} &   {\{64, 32, 32, 16\}}\\
Feature Grouping Attention Heads & 6 & 6 & 6 &   {6} \\
Feature Ungrouping Attention Heads & 6 & 6 & 6 &   {6}\\
MLPMixer Patch Expansion & 0.5 & 0.5 & 0.5 &   {0.5}\\
MLPMixer Channel Expansion & 4 & 4 & 4 &   {4}\\
ImageNet Drop Path Rate & 0.2 & 0.2 & 0.3 &   {0.3}\\
Parameters (M) & 9.3 & 23.6 & 36.2 &   {75.4}\\
\bottomrule
\end{tabular}
\end{adjustbox}
\end{table}

\subsection{Training Recipe for ImageNet}
\vspace{-0.1in}
The ImageNet experiments are based on the MMClassification toolkit~\citep{2020mmclassification}. The models are trained for 300 epochs with a batch size of 2048; the AdamW optimizer was used with a weight decay of 0.05 and a peak learning rate of 0.002. The cosine learning rate schedule is adopted. The gradient clip is set to 5.0 (we also tested 1.0 and found it worked well too); data augmentation strategies are from \cite{Swin_Liu_2021tq} and include Mixup~\citep{zhang2017mixup}, Cutmix~\citep{yun2019cutmix}, Random erasing~\citep{zhong2020random} and Rand augment~\citep{cubuk2020randaugment}.

\subsection{Training Recipe for COCO}
\vspace{-0.1in}
The COCO experiments are based on the MMDetection toolkit~\citep{mmdetection}. Following commonly used training settings, both Mask R-CNN and RetinaNet models are trained for 12 epochs (1$\times$) and 36 epochs (3$\times$). For the 3$\times$ schedule, we follow previous work~\citep{Swin_Liu_2021tq, stunned} to use multi-scale inputs during training. The AdamW optimizer was used with an initial learning rate of 0.0002 and weight decay of 0.05. We used ViTAdapter~\citep{vitadapter} to generate multi-scale features and followed the default hyper-parameter settings in \cite{vitadapter}. 

\subsection{Training Recipe for ADE20K}
\vspace{-0.1in}
The ADE20K experiments are based on the MMSegmentation toolkit~\citep{mmseg2020}. Following commonly used training settings, both UperNet and SegFormer models are trained for 160000 iterations. The input images are cropped to 512$\times$512 during training. The AdamW optimizer was used with an initial learning rate of 0.00006 and weight decay of 0.01. We did not use ViTAdapter~\citep{vitadapter} for segmentation experiments. 

\begin{figure*}[!t]
    \centering
    \vspace{-0.3cm}
    \includegraphics[width=\textwidth]{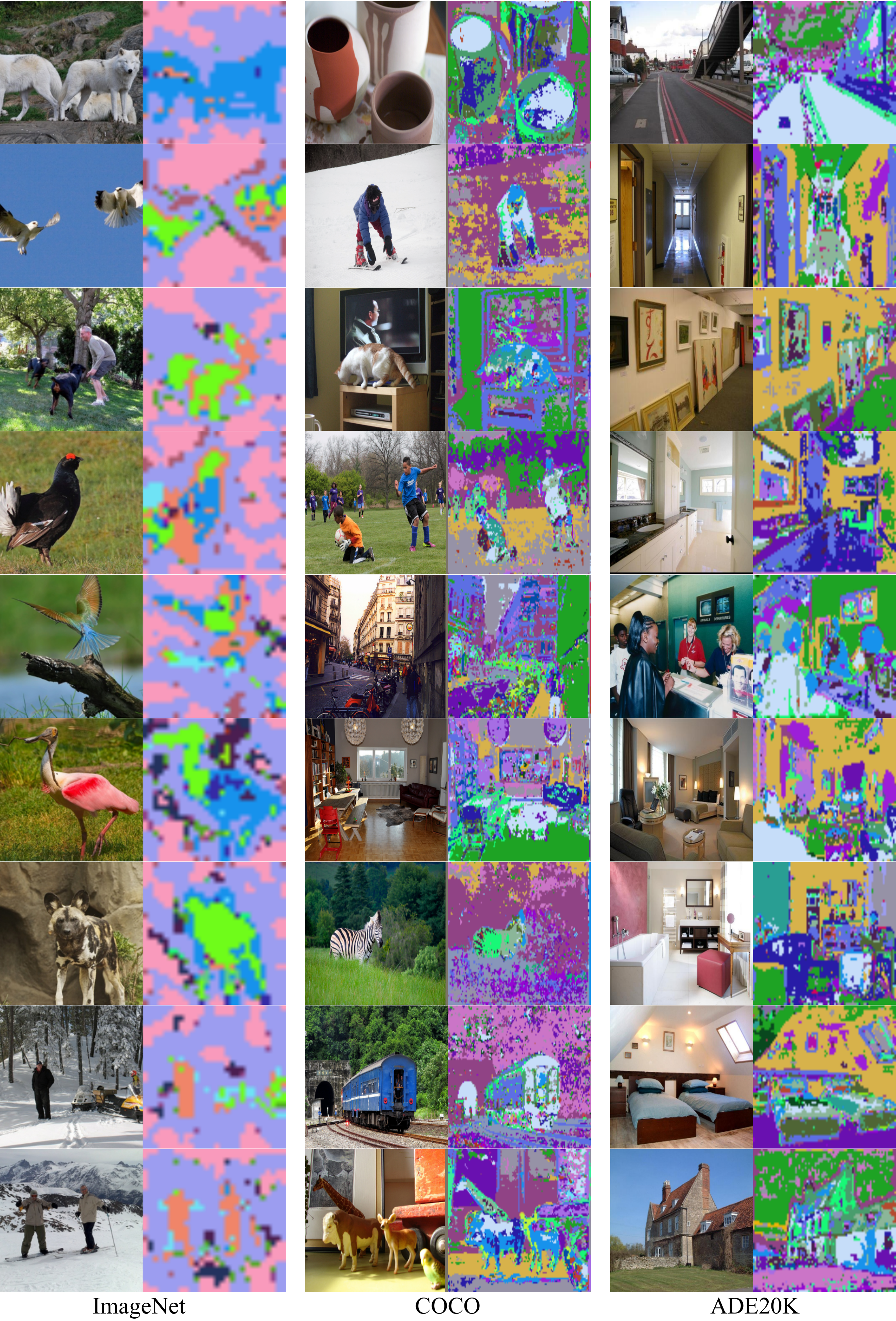}
    \captionsetup{font={small}}
    \vspace{-0.7cm}
    \caption{Feature grouping visualisation using model trained on ImageNet-1k, COCO and ADE20K.}
\label{fig:vis}
\vspace{-0.6cm}
\end{figure*}

\vspace{-0.1in}
\section{Visualizations}
\vspace{-0.1in}
In Figure~\ref{fig:vis}, we visualise the feature grouping results using models trained on ImageNet, COCO and ADE20K. We observe that the feature grouping can separate a image's foreground and background in all three datasets. When the model receives fine-grained supervision like bounding boxes and semantic masks, the feature grouping can correspond to more details in the image.

\section{Comprehensive comparison}
In Table~\ref{table:imagenetFullLarge} and Table~\ref{table:cocoFullLarge}, we provide a more comprehensive comparison between \ourmethod~and other visual recognition models on ImageNet-1k classification and COCO Mask R-CNN object detection and instance segmentation.

\begin{table}[!h]
\centering
\vspace{2mm}
\caption{\small Comprehensive comparison between \ourmethod~and the recent proposed models on ImageNet-1K.}
\label{table:imagenetFullLarge}
\begin{adjustbox}{max width=0.75\linewidth}
\tablestyle{2pt}{0.95}
\begin{tabular}{l@{\hskip -5pt}ccc}
\toprule
Model &  Params (M) & FLOPs (G)  & Top-1 Acc   \\
\midrule
\multicolumn{4}{l}{Hierarchical} \\
\midrule
PVT-T~\citep{PVT_wang2021} &    13.2 &     1.9 &  75.1 \\
PVT-S~\citep{PVT_wang2021} &    24.5 &     3.8 &  79.8 \\
PVT-L~\citep{PVT_wang2021} &    64.1 &     9.8 &  81.7 \\
CvT-13~\citep{CvT_Wu_2021tw} &    20.0 &     4.5 &  81.6 \\
Focal-Tiny~\citep{focal} & 29.1 & 4.9 & 82.2 \\
Focal-Base~\citep{focal} & 89.8 & 16.0 & 83.8 \\
Swin-T~\citep{Swin_Liu_2021tq} &    29 &     4.5 &  81.3 \\
Swin-S~\citep{Swin_Liu_2021tq} &    50 &     8.7 &  83.0 \\
Swin-B~\citep{Swin_Liu_2021tq} &    88 &    15.4 &  83.5 \\
ShiftViT-T~\citep{shiftvit} & 29 &   4.5 & 81.3 \\
ShiftViT-B~\citep{shiftvit} & 89 &   15.6 & 83.3 \\
CSwin-T~\citep{cswin} &    23 &     4.3 &  82.7 \\
CSwin-B~\citep{cswin} &    78 &     15.0 &  84.2 \\
RegionViT-Ti~\citep{chen2022regionvit} & 13.8 &   2.4 & 80.4 \\ 
RegionViT-S~\citep{chen2022regionvit} &    30.6 &     5.3 &  82.6 \\
RegionViT-B~\citep{chen2022regionvit} &    72.7 &     13.0 &  83.2 \\
MViTv2-T~\citep{mvitv2} & 24.0 &  4.7 & 82.3 \\
GC ViT-XXT~\citep{gcvit} & 12.0 &   2.1 & 79.8 \\
GC ViT-XT ~\citep{gcvit} & 20.0 &   2.6 & 82.0 \\
GC ViT-T ~\citep{gcvit} & 28.0 &   4.7 & 83.4 \\
GC ViT-S ~\citep{gcvit} & 51 &   8.5 & 83.9 \\
GC ViT-B ~\citep{gcvit} & 90 &   14.8 & 84.4 \\
DaViT-Tiny~\citep{DAViT} & 28.3 &   4.5 & 82.8 \\
MPViT-S~\citep{MPViT} & 22.8 &   4.7 & 83.0 \\
Shunted-T\citep{stunned} & 11.5 &   2.1 & 79.8 \\
Shunted-S\citep{stunned} & 22.4 &  4.9 & 82.9 \\
DW-T~\citep{dwvit} & 30.0 &   5.2 & 82.0 \\
DW-B~\citep{dwvit} & 91.0 &   17.0 & 83.8 \\
DAT-T~\citep{DAT} & 29.0 &   4.6 & 82.0 \\
DAT-B~\citep{DAT} & 88.0 &   15.8 & 84.0 \\
\midrule
\multicolumn{4}{l}{Non-hierarchical} \\
\midrule
DeiT-S~\citep{DeiT_touvron2020} &    22.1 &     4.6 &  79.9 \\
DeiT-B~\citep{DeiT_touvron2020} &    86 &     16.8 &  81.8 \\
CaiT-XS24~\citep{CaiT_Touvron:2021tp} & 26.6 & 5.4 & 81.8 \\
CaiT-S48~\citep{CaiT_Touvron:2021tp} & 89.5 & 18.6 & 83.5 \\
LocalViT-S~\citep{LocalViT_Li_2021ww} &    22.4 &     4.6 &  80.8 \\
Visformer-S~\citep{Visformer_Chen_2021ve} & 40.2 &   4.9 & 82.3 \\
ConViT-S~\citep{ConViTdAscoli_2021vz} &    27.0 &     5.4 &  81.3 \\
ConViT-B~\citep{ConViTdAscoli_2021vz} &    86.0 &     17.0 &  82.4\\
\midrule
\ourmethod-L1 & 9.3 &   5.8 & 80.5 \\
\ourmethod-L2 & 23.8 &   15.0 & 83.4 \\
\ourmethod-L3 & 36.2 &   22.9 & 84.1 \\
   {\ourmethod-L4} &    {75.4} &      {48.2} &    {84.3} \\
\bottomrule
\end{tabular}
\end{adjustbox}
\vspace{-1mm}

\end{table}

\begin{table}[!h]
\centering
\caption{\small Comprehensive comparison of Mask R-CNN object detection and instance segmentation on MS COCO \textit{mini-val} using 1$\times$ and 3$\times$ + MS schedule.}
\label{table:cocoFullLarge}
\vspace{-3mm}
\begin{adjustbox}{max width=0.8\linewidth}
\tablestyle{2pt}{0.95}
\begin{tabular}{lcc|cc|cc}
\toprule
         &      Params      &    FLOPs       & \multicolumn{2}{c|}{1$\times$} & \multicolumn{2}{c}{3$\times$} \\
Backbone &  (M) &  (G) & $AP^{bb}$  & $AP^{mk}$ & $AP^{bb}$ & $AP^{mk}$  \\

\midrule
\multicolumn{7}{l}{Hierarchical} \\
\midrule  
ResNet-50 ~\citep{ResNet_He_2016_CVPR} & 44 & 260  & 38.0 & 34.4 & 41.0 & 47.1 \\
ResNet101~\citep{ResNet_He_2016_CVPR} & 63 & 336  & 40.4 & 36.4  &  42.8 & 38.5 \\
ResNeXt101-32x4d~\citep{ResNeXt_Xie_2017_CVPR} & 62 & 340 & 41.9 & 37.5 &  44.0 & 39.2  \\
RegionViT-S~\citep{chen2022regionvit} & 50.1 & 171 & 42.5 & 39.5 & 46.3 & 42.3  \\
RegionViT-B~\citep{chen2022regionvit} & 92 & 287 & 44.2 & 40.8 & 47.6 & 43.4  \\
PVT-Tiny~\citep{PVT_wang2021} & 32.9 & - & 36.7 & 35.1 &  39.8 & 37.4 \\
PVT-Small~\citep{PVT_wang2021} & 44.1 & - & 40.4 & 37.8 &  43.0 & 39.9 \\
PVT-Medium~\citep{PVT_wang2021} & 63.9 & - & 42.0 & 39.0 &  44.2 & 40.5 \\
PVT-Large~\citep{PVT_wang2021} & 81.0 & 364 & 42.9 & 39.5 &  44.5 & 40.7 \\
Swin-S~\citep{Swin_Liu_2021tq} & 69 & 354 & 44.8 & 40.9 & 47.6 & 42.8 \\
Swin-B~\citep{Swin_Liu_2021tq} & 107 & 496 & 45.5 & 41.3 & - & - \\
MViTv2-T~\citep{mvitv2} & 44 & 279 & - & - & 48.2 & 43.8 \\
MViTv2-S~\citep{mvitv2} & 54 & 326 & - & - & 49.9 & 45.1 \\
MViTv2-B~\citep{mvitv2} & 71 & 392 & - & - & 51.0 & 45.7 \\
DaViT-Small~\citep{DAViT} & 69 & 351 & 47.7 & 42.9 & 49.5 & 44.3\\
DaViT-Base~\citep{DAViT} & 107 & 491 & 48.2 & 43.3 & 49.9 & 44.6\\
DAT-T~\citep{DAT} & 48 & 272 & 44.4 & 40.4 & 47.1 & 42.4 \\
DAT-S~\citep{DAT} & 69 & 387 & 47.1 & 42.5 & 49.0 & 44.0 \\
DW-B~\citep{dwvit} & 111 & 505 & - & - & 49.2 & 44.0 \\
CMT-S~\citep{CMT} & 44 & 231 & 44.6 & 40.7 & - & - \\
CSwin-S~\citep{cswin} &    54 &    342 &  47.9 & 43.2 & 50.0 & 44.5 \\
CSwin-B~\citep{cswin} &    97 &    526 &  48.7 & 43.9 & 50.8 & 44.9 \\
MPViT-B~\citep{MPViT} &    28 &    216 &  - & - & 44.8 & 41.0 \\
MPViT-B~\citep{MPViT} &    30 &    231 &  - & - & 46.6 & 46.1 \\
MPViT-B~\citep{MPViT} &    43 &    268 &  - & - & 48.4 & 47.6 \\
MPViT-B~\citep{MPViT} &    95 &    503 &  - & - & 49.5 & 44.5 \\
Shunted-S\citep{stunned} & 42 & - &47.1 & 52.1 & 49.1 & 43.9 \\
Shunted-B\citep{stunned} & 59 & - &48.0 & 43.2 & 50.1 & 45.2 \\
GC ViT-T~\citep{gcvit} & 47 & 263 & - & -  & 46.5 & 41.8 \\
\midrule
\multicolumn{7}{l}{Non-hierarchical} \\
\midrule
ViT-Adapter-T~\citep{vitadapter} & 28 & - & 41.1 & 37.5 & 46.0 & 41.0 \\
ViT-Adapter-S~\citep{vitadapter} & 47 & - & 44.7 & 39.9 & 48.2 & 42.8 \\
ViT-Adapter-B~\citep{vitadapter} & 102 & - & 47.0 & 41.8 & 49.6 & 43.6 \\
\midrule
\ourmethod-L1 & 33 & 457 & 48.1 & 42.7 & 50.2 & 44.3 \\
\ourmethod-L2 & 50 & 690 & 49.9 & 43.9 & 51.4 & 45.1 \\
\ourmethod-L3 & 64 & 884 & 50.4 & 44.4 & 51.6 & 45.2 \\
    {\ourmethod-L4} &     {109} &     {1489} &     {51.0} &     {45.0} &     {52.1} &     {45.7} \\
\bottomrule
\end{tabular}
\end{adjustbox}
\vspace{-6mm}

\end{table}

\end{document}